\documentclass[sigconf]{acmart}

\usepackage{booktabs} 
\usepackage{multirow}
\usepackage{graphicx, caption, subcaption}
\usepackage{color, colortbl}
\definecolor{Gray}{gray}{0.9}
\usepackage{mathtools}
\usepackage{footmisc}
\usepackage{balance} 
\usepackage{flushend}

\def\x{{\mathbf x}}
\pdfoutput=1

\def \reducevspace {-0.7em}

\copyrightyear{2017}
\acmYear{2017}
\setcopyright{acmcopyright}
\acmConference{MM'17}{}{October 23--27, 2017, Mountain View, CA, USA.}
\acmPrice{15.00}
\acmDOI{https://doi.org/10.1145/3123266.3123417}
\acmISBN{978-1-4503-4906-2/17/10}

\begin{document}
\title{Selective Deep Convolutional Features for Image Retrieval}

\author{Tuan Hoang}
\affiliation{%
  \institution{Singapore University of Technology and Design}
}
\email{nguyenanhtuan_hoang@mymail.sutd.edu.sg}

\author{Thanh-Toan Do}
\affiliation{%
  \institution{The University of Adelaide}
}
\email{thanh-toan.do@adelaide.edu.au}

\author{Dang-Khoa Le Tan}
\affiliation{%
  \institution{Singapore University of Technology and Design}
}
\email{letandang_khoa@sutd.edu.sg}

\author{Ngai-Man Cheung}
\affiliation{
  \institution{Singapore University of Technology and Design}
}
\email{ngaiman_cheung@sutd.edu.sg}

\begin{abstract}
Convolutional Neural Network (CNN) is a very powerful approach to extract discriminative local descriptors for effective image search. Recent work adopts fine-tuned strategies to further improve the discriminative power of the descriptors. Taking a different approach, in this paper, we propose a novel framework to achieve competitive retrieval performance. 
Firstly, we propose 
various masking schemes, namely \textbf{\textit{SIFT-mask}}, \textbf{\textit{SUM-mask}}, and \textbf{\textit{MAX-mask}}, to select a representative subset of local convolutional features and remove a large number of redundant features. 
We demonstrate that this can effectively address the burstiness issue and improve retrieval accuracy. 
Secondly, we propose to employ recent embedding and aggregating methods to further enhance feature discriminability. Extensive experiments demonstrate that our proposed framework achieves state-of-the-art retrieval accuracy. 
\end{abstract}

%
%



\maketitle

\section{Introduction}
Content-based image retrieval (CBIR) has attracted a sustained attention from the multimedia/computer vision community due to its wide range of applications, e.g. visual search, place recognition. Earlier works heavily rely on hand-crafted local descriptors, e.g. SIFT \cite{SIFT_Lowe} and its variant \cite{rootsift}. Even though there are great improvements of the SIFT-based image search systems over time, the performance of these systems still has room for improvement. There are two main issues: the first and the most important one is that SIFT features lack discriminability \cite{cnn_max_pooling} to emphasize the differences in images. Even though this drawback is  relieved to some extent when embedding local features to much higher dimensional space \cite{bow,FisherVector,vlad,Temb,FAemb,F-FAemb}, there is still a large semantic gap between SIFT-based image representation and human perception on instances (objects/scenes) \cite{cnn_max_pooling}. Secondly, the strong effect of {\em   burstiness} \cite{burstiness}, i.e. numerous descriptors are almost similar within the same image, considerably degrade the quality of SIFT-based image representation for the image retrieval task \cite{burstiness,vlad,revisitvlad}.

Recently, deep Convolutional Neural Networks (CNN) have achieved a lot of success in various problems including image classification \cite{Alexnet,VGG,googlenet,resnet}, object detection \cite{RCNN,FasterRCNN}, etc. After training a CNN on a huge annotated dataset, e.g. ImageNet \cite{ILSVRC15}, 
outputs of middle/last layers can capture rich information at higher semantic levels. 
On one hand, the output of the deeper layer possesses abstract understanding of images  for  computer vision tasks that require high-invariance to the intra-class variability, e.g.,  classification, detection \cite{Alexnet,VGG,googlenet,resnet,RCNN,FasterRCNN}. On the other hand, the middle layers contain more visual information on edges, corners, patterns, and structures.  Therefore, they are more suitable for image retrieval \cite{cnn_max_pooling,CroW,R-MAC,netvlad,MOF}.
Utilizing the outputs of the convolutional layers to produce the image representation, recent image retrieval methods \cite{cnn_max_pooling,CroW,R-MAC,netvlad,MOF} 
achieve a considerable performance boost. 

Although the local convolutional (conv.) features are more discriminative than SIFT features \cite{cnn_max_pooling}, to the best of our knowledge, none of the previous works has considered the burstiness problem
which appears in the local features. 
In this paper, focusing on CNN based image retrieval, we delve deeper into the issue: \textit{``How to eliminate redundant local features in a robust way?''} Since elimination of redundant local features leads to better representation and faster computation, we emphasize both aspects in our  experiments. Specifically, 
inspired by the concept of finding a set of interest regions before deriving their corresponding local features - the concept which has been used in design of hand-crafted features, we propose three different masking schemes for selecting \textit{representative} local conv. features, including \textbf{\textit{SIFT-mask}}, \textbf{\textit{SUM-mask}}, and \textbf{\textit{MAX-mask}}. The principal ideas of \textbf{our main contribution} are that we take advantages of SIFT detector \cite{SIFT_Lowe} to produce SIFT-mask; moreover, we utilize sum-pooling and max-pooling over all conv. feature channels to derive SUM-mask and MAX-mask, respectively.

Additionally, most of the recent works which take local conv. features as input \cite{R-MAC,CroW,finetune_hard_samples} do not
leverage local feature embedding and aggregating \cite{FisherVector,vlad,Temb,FAemb}, which are effective processes to enhance the discriminability for hand-crafted features. In \cite{cnn_max_pooling}, the authors mentioned that the deep convolutional features are already discriminative enough for image retrieval task, hence, the embedding is not necessary to enhance their discriminability. However, in this work, we find that  by utilizing state-of-art embedding methods on our selected deep convolutional features \cite{Temb,FAemb,vlad,FisherVector}, we can significantly   enhance the discriminability. 
Our experiments show that applying embedding and aggregating on our selected local conv. features significantly improves image retrieval accuracy.

The remaining of this paper is organized as follows. Section \ref{sec:related-work} discusses related works. Section \ref{sec:selective_local_fea} presents the details of our main contribution, the masking schemes, together with preliminary experimental results to justify their effectiveness. In section \ref{sec:frame}, we introduce the proposed framework for computing the final image representation which takes selected local deep conv. features as input and output a global fixed-length image representation. Section \ref{sec:exp} presents a wide range of experiments to comprehensively evaluate our proposed framework.
Section \ref{sec:conclusion} concludes the paper.

\vspace{\reducevspace}
\section{Related work}
\label{sec:related-work}
In the task of image retrieval, the early CNN-based work  \cite{neuralcode} takes the activation of fully connected layers as global descriptors followed by dimensionality reduction. This work shows that supervised retraining the network on the dataset which is relevant to the test set is very beneficial in the retrieval task. 
However, as shown in \cite{neuralcode}, the creation of labeled training data is expensive and non-trivial. 
\citet{MOP} proposed Multi-Scale Orderless Pooling (MOP)  to embed and pool the CNN fully-connected activations of image patches of an image at different scale levels. This enhances the scale invariant of the extracted features. However, the method is computationally expensive because multiple patches (resized to the same size) of an image are fed forward into the CNN. The recent work \cite{CNN-vs-SIFT} suggested that CNN fully-connected activations and SIFT features are highly complementary. They proposed to integrate SIFT features with fully-connected CNN features at different levels.

Later works shift the focus from fully-connected layers to conv. layers for extracting image features because lower layers are more general and certain level of spatial information is still preserved \cite{generic2specific}. When conv. layers are used, the conv. features are usually considered as local features, hence, a pooling method (sum or max) is   applied on the conv. features to produce the single image representation.
\citet{cnn_max_pooling} showed that sum-pooling outperforms max-pooling when the final image representation is whitened. 
\citet{CroW} further improved sum-pooling on conv. features by proposing a non-parametric method to learn weights for both spatial locations and feature channels. \citet{R-MAC} revisited max-pooling by proposing the strategy to aggregate the maximum activation over multiple spatial regions sampled on the final conv. layer using a \textit{fixed layout}. This work together with \cite{CroW} are currently the state-of-art methods in image retrieval task using off-the-shelf CNN.

Although  fine-tuning an off-the-shelf network (e.g. AlexNet or VGG) can enhance the discriminability of the deep features \cite{neuralcode} for image retrieval, the collecting of training data is non-trivial. Recent works tried to overcome this challenge by proposing unsupervised/weak supervised fine-tuning approaches which are specific for image retrieval.  
\citet{netvlad} proposed a new generalized VLAD layer and this layer can be stacked with any CNN architecture. The whole architecture, named NetVLAD, is trained in an end-to-end manner in which the data is collected in a weakly supervised manner from Google Street View Time Machine. Also taking the approach of fine-tuning the network in a weakly-supervised manner, \citet{quartet-net} proposed an automatic method to harvest data from GeoPair dataset \cite{goepair} to train a special architecture called Quartet-net with the novel double margin contrastive loss function. 
 Concurrently, \citeauthor{finetune_hard_samples} \cite{finetune_hard_samples}  proposed a different approach to re-train state-of-the-art CNNs of classification task for image retrieval. They take advantages of 3D reconstruction to obtain matching/non-matching pairs of images in an unsupervised manner for re-training process. 

\section{Selective Local Deep Conv. Features}
\label{sec:selective_local_fea}

\def \imagescale {0.145}
\begin{figure*}[t]
\centering
\begin{subfigure}[b]{\imagescale\textwidth}
\frame{\includegraphics[width=\textwidth]{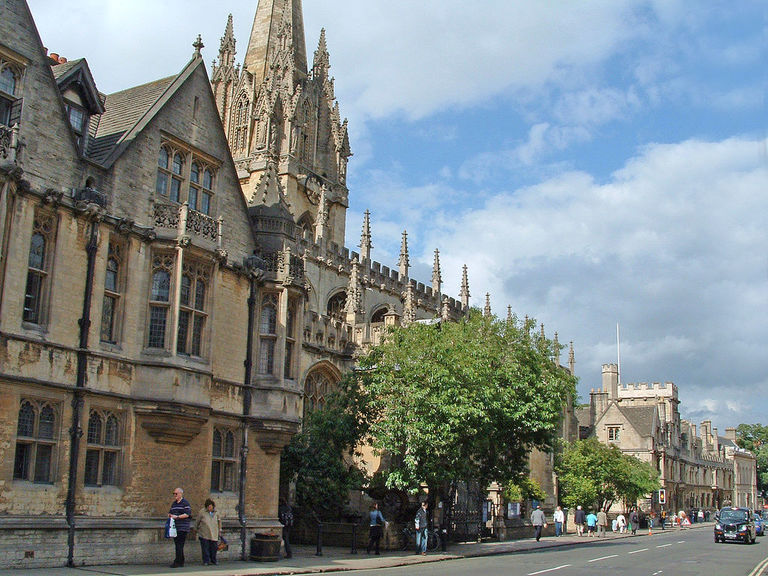}}
\end{subfigure}
\quad
\begin{subfigure}[b]{\imagescale\textwidth}
\frame{\includegraphics[width=\textwidth]{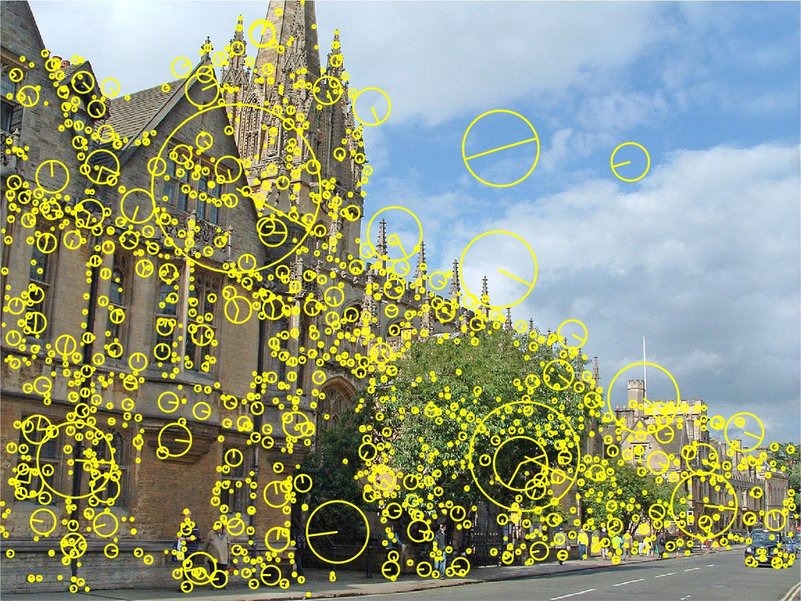}}
\end{subfigure}
\quad
\begin{subfigure}[b]{\imagescale\textwidth}
\frame{\includegraphics[width=\textwidth]{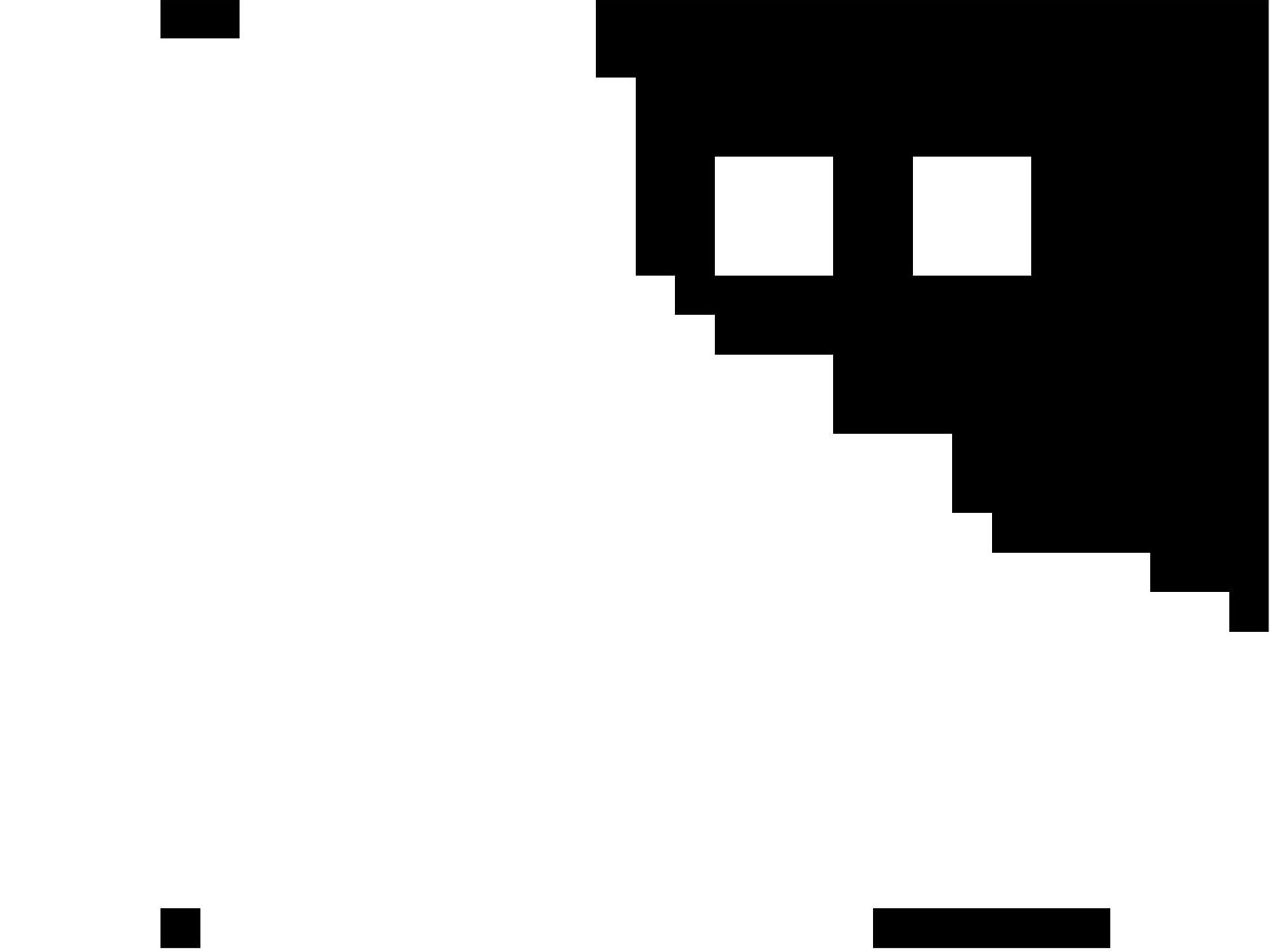}}
\end{subfigure}
\quad
\begin{subfigure}[b]{\imagescale\textwidth}
\frame{\includegraphics[width=\textwidth]{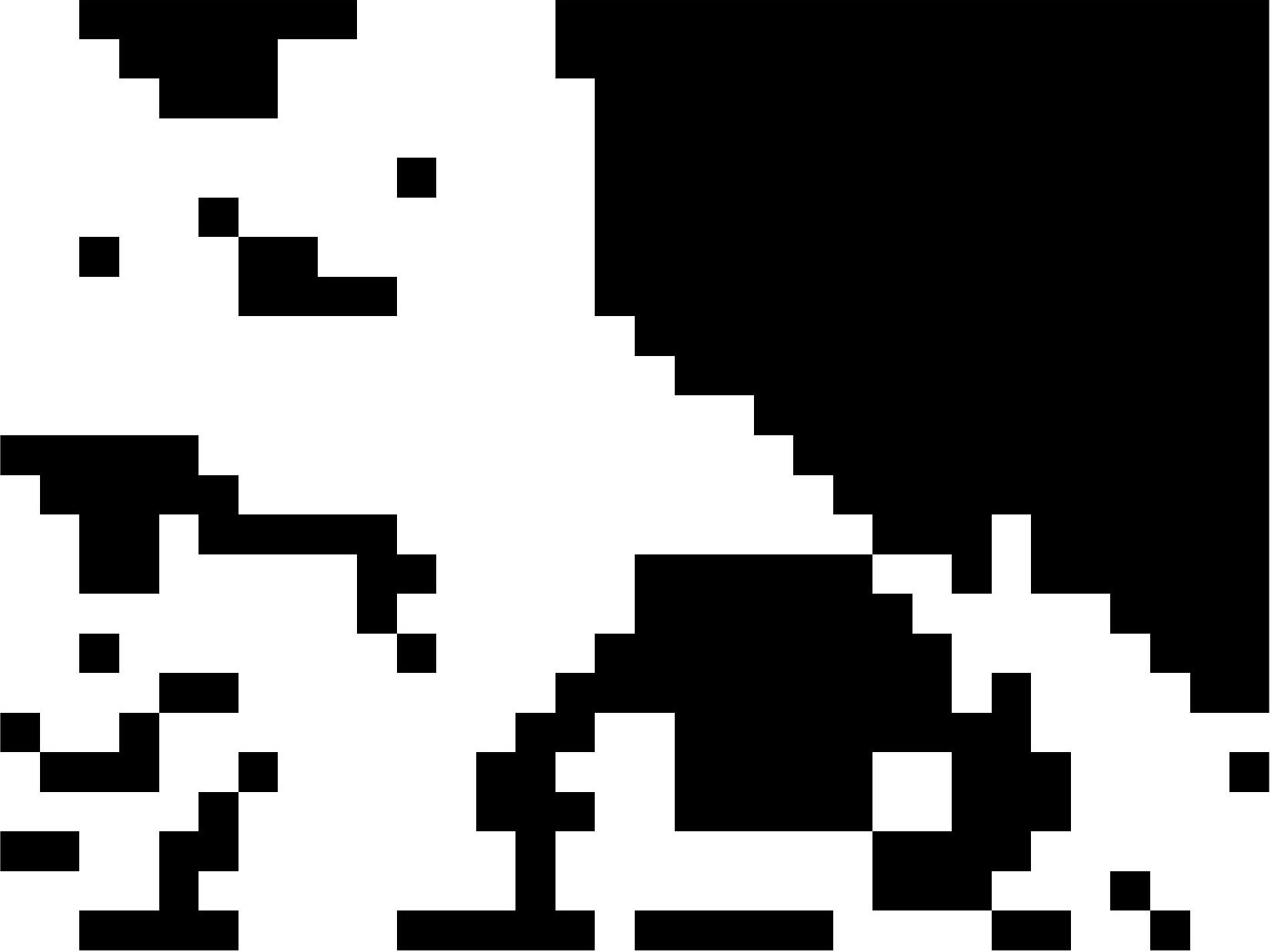}}
\end{subfigure}
\quad
\begin{subfigure}[b]{\imagescale\textwidth}
\frame{\includegraphics[width=\textwidth]{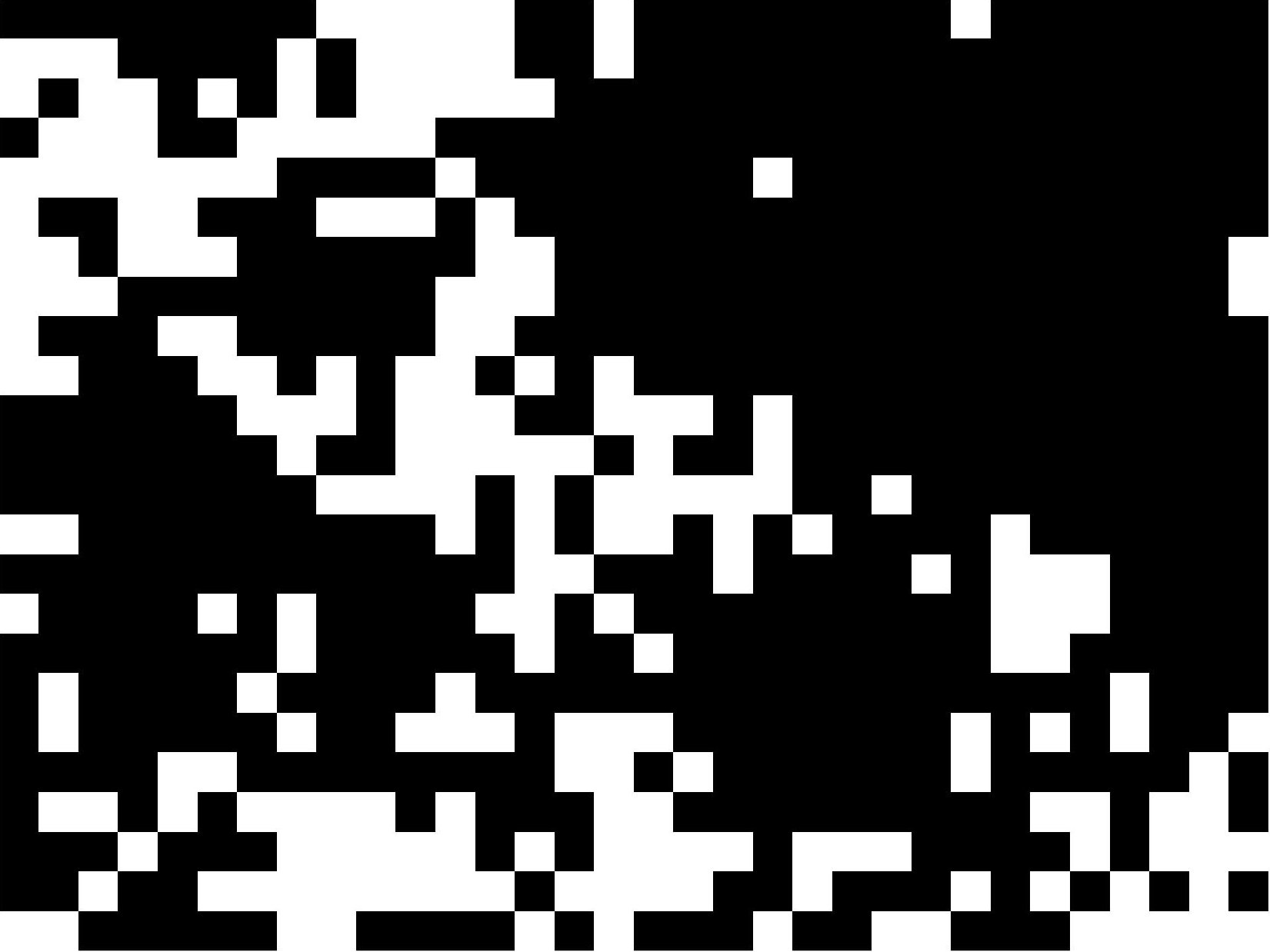}}
\end{subfigure}

\begin{subfigure}[b]{\imagescale\textwidth}
\frame{\includegraphics[width=\textwidth]{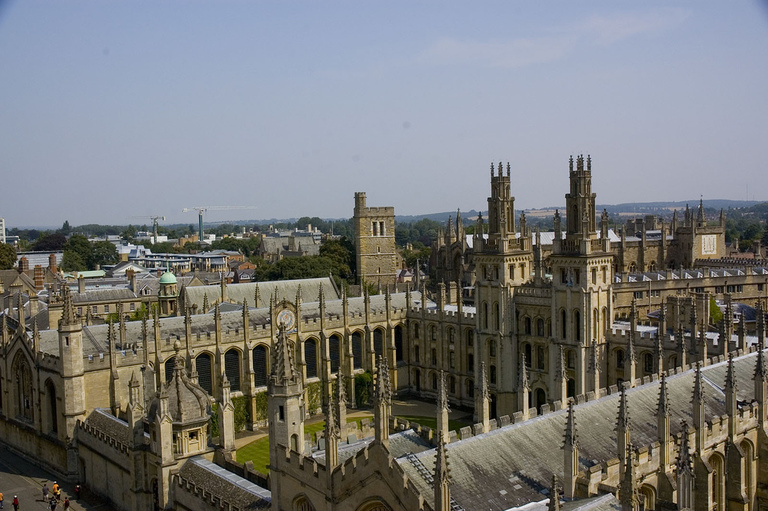}}
\caption{Original image}
\label{sfig:orig_img}
\end{subfigure}
\quad
\begin{subfigure}[b]{\imagescale\textwidth}
\frame{\includegraphics[width=\textwidth]{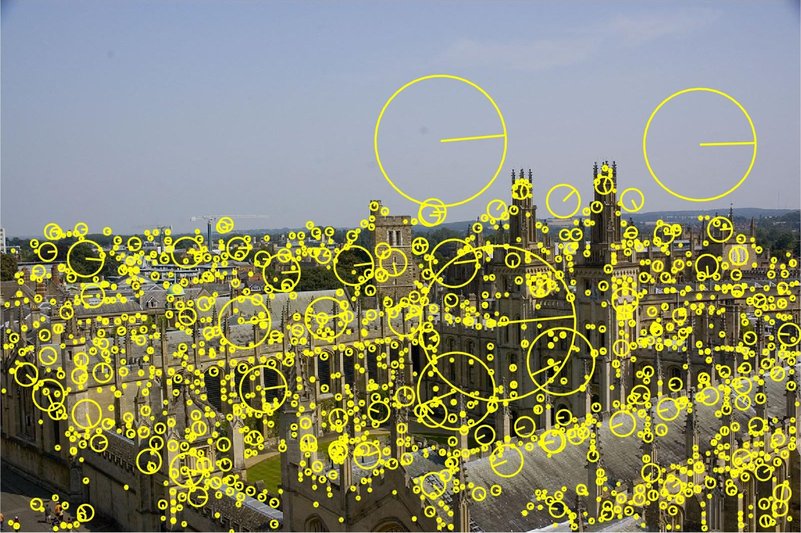}}
\caption{Image+SIFT}
\label{sfig:img+sift}
\end{subfigure}
\quad
\begin{subfigure}[b]{\imagescale\textwidth}
\frame{\includegraphics[width=\textwidth]{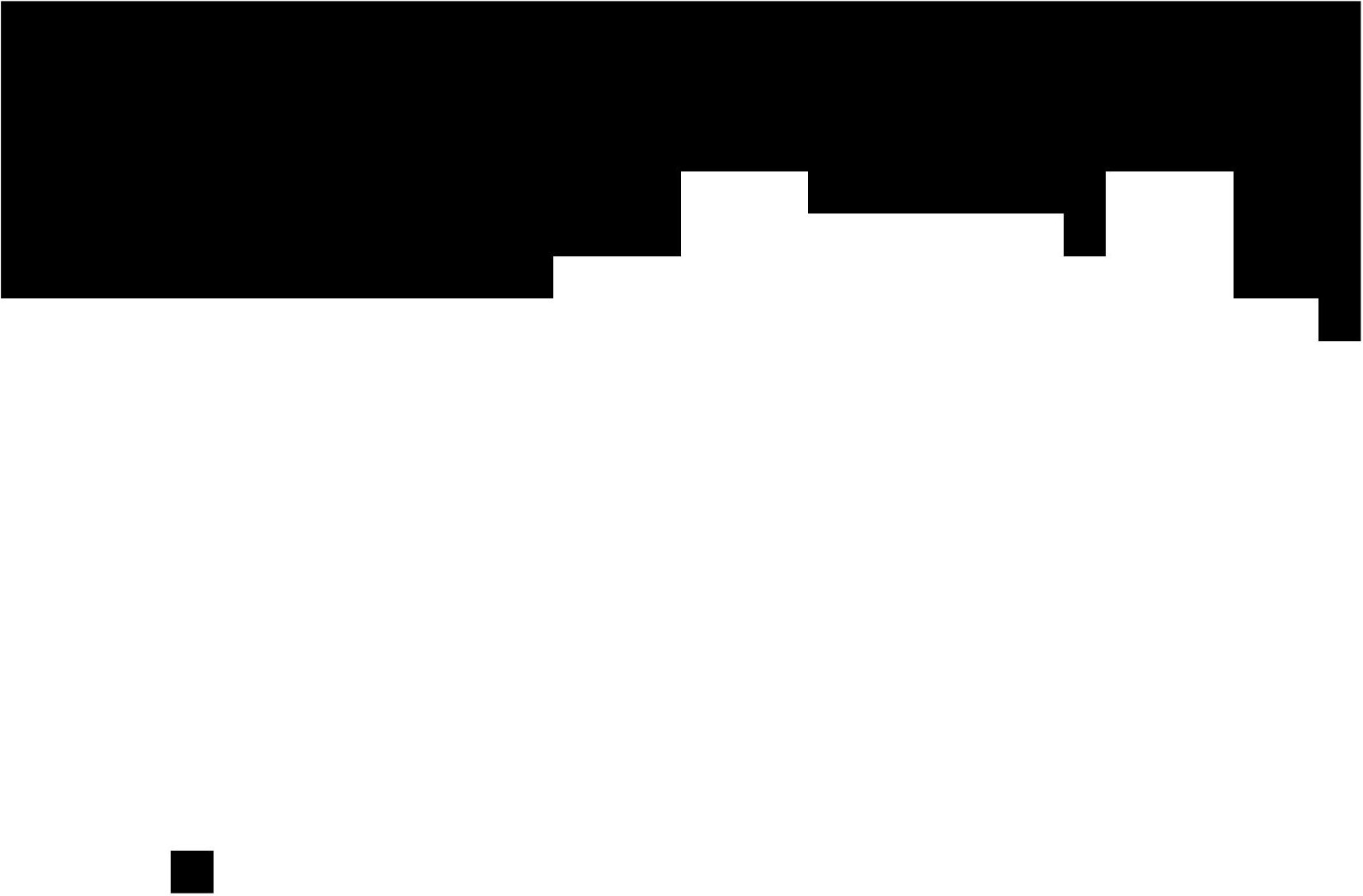}}
\caption{SIFT-mask}
\label{sfig:sift_mask}
\end{subfigure}
\quad
\begin{subfigure}[b]{\imagescale\textwidth}
\frame{\includegraphics[width=\textwidth]{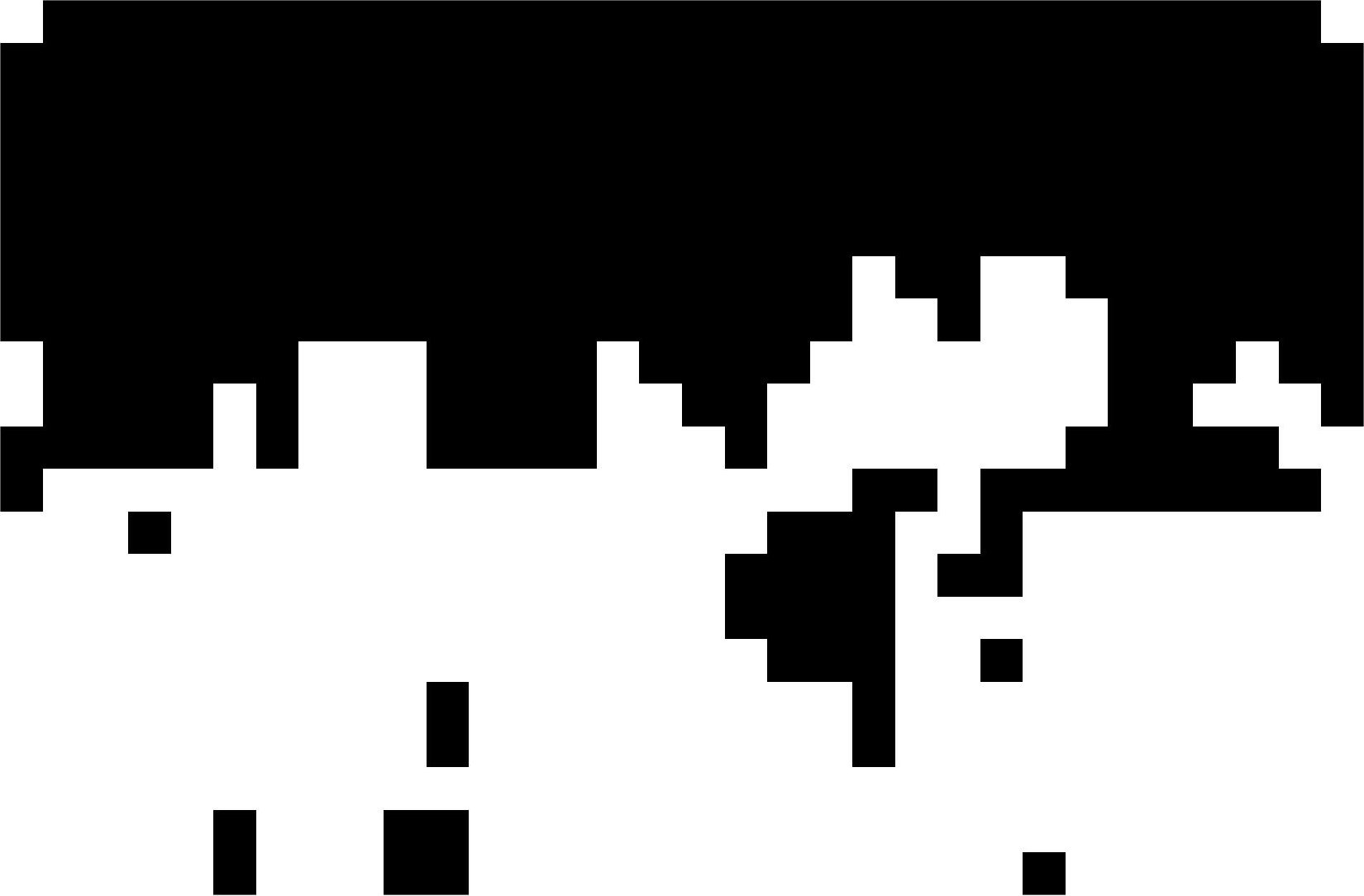}}
\caption{SUM-mask}
\label{sfig:sum_mask}
\end{subfigure}
\quad
\begin{subfigure}[b]{\imagescale\textwidth}
\frame{\includegraphics[width=\textwidth]{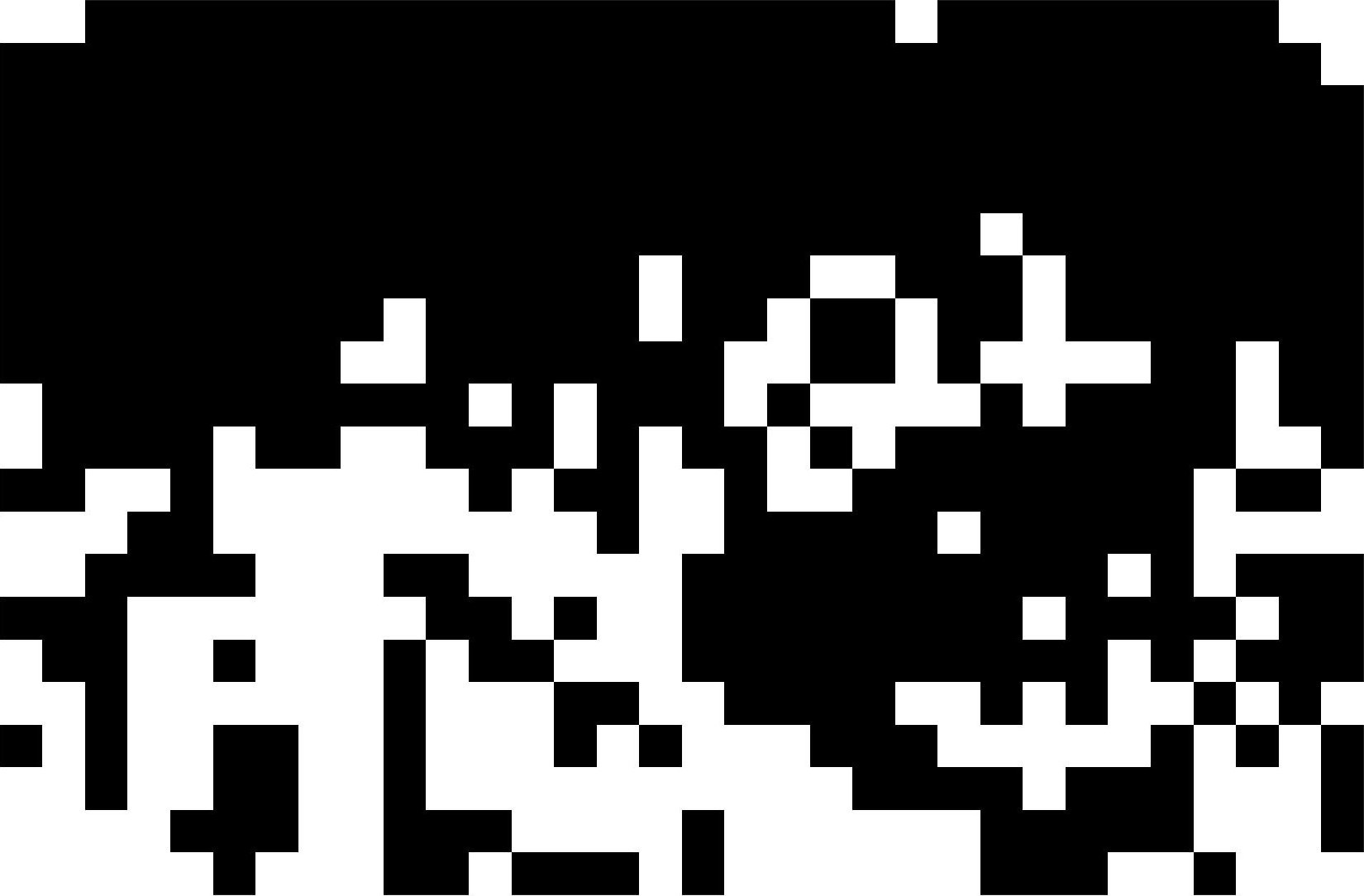}}
\caption{MAX-mask}
\label{sfig:max_mask}
\end{subfigure}

\caption{Examples of masks to select local features. The original images are showed on the first column (\ref{sfig:orig_img}). The second column shows regions which are covered by SIFT features. The SIFT/SUM/MAX-masks of corresponding images are showed in the last three columns (\ref{sfig:sift_mask},\ref{sfig:sum_mask},\ref{sfig:max_mask}).}
\label{fig:mask-example}
\end{figure*}

In this section, we first define the set of local deep conv. features which we work on throughout the paper (Section \ref{ssec:local_features}). We then present proposed strategies for selecting a subset of discriminative local conv. features, including \textbf{\textit{SIFT mask}}, \textbf{\textit{SUM mask}}, and \textbf{MAX mask} (Section \ref{ssec:selective_fea}).
Finally, we discuss experiments to illustrate the effectiveness of our methods (Section \ref{ssec:effectiveness}). 

\subsection{Local deep convolutional features}
\label{ssec:local_features}
We consider a pre-trained CNN with all the fully connected layers discarded. Given an input image $I$ of size $W_I \times H_I$ that is fed through a CNN, the 3D activations (responses) tensor of a conv. layer has the size of $W \times H \times K$ dimensions, where $K$ is the number feature maps and $W \times H$ is the spatial resolution of a feature map. 
We consider this 3D tensor of responses as a set of $(W \times H)$ local features; each of them have $K$ dimensions. In other words, each position on the $W \times H$ spatial grid  is the location of a local feature. Each local conv. feature is a vector of $K$ values of the $K$ feature maps at a particular location. We denote $\mathcal{F}^{(k)}$ as $k^{th}$ feature map (and its size is $W \times H$). 
Note that the choice of the conv. layer to be used is not fixed in our method. We investigate the impact of choosing different conv. layers in Section \ref{sec:exp}.

\subsection{Selective features}
\label{ssec:selective_fea}
We now formally propose different methods to compute a selection mask, i.e. a set of unique coordinates $\{(x,y)\}$ in the feature maps where local conv. features are retained $(1 \le x \le W;1 \le y \le H)$.
Our proposed methods for selecting discriminative local deep conv. features are inspired by the concept of finding the interest regions in the input images which is traditionally used in the design of hand-crafted features. 

\subsubsection{SIFT Mask.} 
\label{sssec:sift_mask}
Prior the emergence of CNN features in the image retrieval task, most previous works \cite{Temb,FAemb, burstiness,vlad,FisherVector,neg_evidence,to_aggregate,revisitvlad} are based on SIFT \cite{SIFT_Lowe} features and its variant RootSIFT \cite{rootsift}. 
Even though it has been showed that there is still a gap between SIFT-based representation and the semantic meaning in the image, these works have clearly demonstrated the capability of SIFT feature, especially in the aspect of key-point detection. 
Figure (\ref{sfig:img+sift}) shows local image regions which are covered by SIFT. 
We can obverse that regions covered by SIFT mainly focus on the salient regions, i.e., buildings. This means that SIFT keypoint detector is capable to locate important regions of images. Hence, we propose a method which takes advantage of SIFT detector in combination with highly-semantic local deep conv. features. 

Specifically, let set $\mathcal{S}=\{(x^{(i)}, y^{(i)})\}_{i=1}^{n}$ be SIFT feature locations extracted from an image with the size of $W_I\times H_I$; each location on the spatial grid $W \times H$ is location of a local deep conv. feature. Based on the fact that convolutional layers still preserve the spatial information of the input image \cite{R-MAC}, we select a subset of locations on the spatial grid which correspond to locations of SIFT key-points, i.e., 
\begin{equation}
\displaystyle{\mathcal{M}_{\text{SIFT}} = \left\{
\left(x_\text{SIFT}^{(i)}, y_\text{SIFT}^{(i)}\right) \right\}
\qquad i=1, \cdots, n }
\end{equation}
where ${x_\text{SIFT}^{(i)}=\text{round}\left( \frac{x^{(i)}W}{W_I}\right)}$ and ${y_\text{SIFT}^{(i)}=\text{round}\left( \frac{y^{(i)}H}{H_I}  \right)}$, in which $\text{round}(\cdot)$ represents rounding to nearest integer. 
By keeping only locations $\mathcal{M}_{\text{SIFT}}$, we expect to remove ``background'' deep conv. features, while keeping ``foreground'' ones.

Note that SIFT detector has the issue of  burstiness \cite{burstiness}.
However, regarding local conv. features, this burstiness effect is expected to be less severe since local conv. features have much larger receptive fields than those of SIFT features. For examples, a local conv. feature from $\mathtt{pool5}$ layers of AlexNet \cite{Alexnet} and VGG16 \cite{VGG} covers a region of $195\times 195$ and $212\times 212$ in the input image, respectively.

\subsubsection{MAX Mask}
\label{sssec:max_mask} 
It is known that each feature map contains the activations of  a specific visual structure \cite{visual_CNN,RCNN}. Hence, we propose to select a subset of local conv. features which contain  high activations for all visual contents, i.e. we select the local features that capture the most prominent structures in the input images. This property, actually, is desirable to distinguish scenes.

Specifically, we assess each feature map and select the location corresponding to the max activation value on that feature map. Formally, we define the selected locations $\mathcal{M}_{\text{MAX}}$ as follows:
\begin{equation}
\begin{split}
&\mathcal{M}_{\text{MAX}} = \left\{\left(x_\text{MAX}^{(k)}, y_\text{MAX}^{(k)}\right)\right\}\qquad k=1, \cdots, K  \\
&\left(x_\text{MAX}^{(k)}, y_\text{MAX}^{(k)}\right) = \arg\max\limits_{(x,y)}\mathcal{F}_{(x,y)}^{(k)}
\end{split}
\end{equation}

\subsubsection{SUM Mask} Departing from the MAX-mask idea, we propose a different masking method based on the idea that a local conv. feature is more \textit{informative} if it gets excited in more feature maps, i.e., the sum on description values of a local feature is larger. 
By selecting local features that have large values of sum, we can expect that those local conv. features contain a lot of information from different local image structures \cite{visual_CNN}.  Formally, we define the selected locations $\mathcal{M}_{\text{SUM}}$ as follows:
\begin{equation}
\begin{split}
&\mathcal{M}_{\text{SUM}} = \left\{(x,y) \;|\; \Sigma_{(x,y)}^\mathcal{F} \ge \alpha \right\}\\
&\Sigma_{(x,y)}^\mathcal{F} = \sum_{k=1}^K\mathcal{F}_{(x,y)}^{(k)} \qquad  \qquad \alpha=\text{median}(\Sigma^\mathcal{F})
\end{split}
\end{equation}


\subsection{Effectiveness of masking schemes}
\label{ssec:effectiveness}
In this section, we evaluate the effectiveness of our proposed masking schemes in eliminating redundant local conv. features. 
Firstly, Figure \ref{sfig:num_local_fea} shows the averaged percentage  of the remaining local conv. features after applying our proposed masks on three datasets: \textit{Oxford5k} \cite{oxford5k}, \textit{Paris6k} \cite{paris6k}, and \textit{Holidays} \cite{holiday}. Clearly, there are a large number of local conv. features removed, about 25\%, 50\%, and 70\% for SIFT/SUM/MAX-mask respectively\footnote{With the input image sizes of $\max(W_I,H_I)=1024.$}.
Additionally, we present the normalized histograms of covariances of selected local conv. features after applying different masks in Figure \ref{sfig:sift_hist}, \ref{sfig:sum_hist}, and \ref{sfig:max_hist}. To compute the covariances, we first $l2$-normalize local conv. features, which are extracted from $\mathtt{pool5}$ layer of the pre-trained VGG \cite{VGG} (the input image is of  size  $\max(W_I,H_I)=1024$). We then compute the dot products for all pairs of features. For  comparison, we include the  normalized histograms of covariances of all available local conv. features (i.e., before masking). 
These figures clearly show that the distributions of covariances after applying masks have much higher peaks around 0 and have smaller tails than those without applying masks.  This indicates some reduction of correlation between the features with the use of mask. Furthermore, Figure \ref{sfig:std} shows the averaged percentage of $l2$-normalized feature pairs that have dot products in the range of $[-0.15, 0.15]$. The chart shows that the selected features are more uncorrelated. 
In summary, Figure \ref{fig:hist} suggests that our proposed masking schemes can help to remove a large proportion of redundant local conv. features, hence achieving a better representative set of local conv. features. Note that with the reduced number of features, we can reduce computational cost, e.g. embedding of features in the subsequent step.

\def \histscale {0.19}
\begin{figure*}[ht]
\centering
\begin{subfigure}[b]{\histscale\textwidth}
\includegraphics[width=\textwidth]{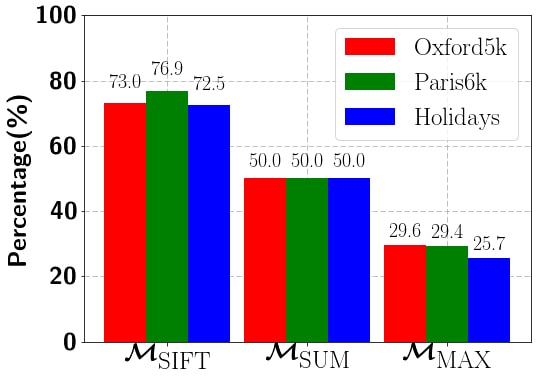}
\caption{}
\label{sfig:num_local_fea}
\end{subfigure}
~
\begin{subfigure}[b]{\histscale\textwidth}
\includegraphics[width=\textwidth]{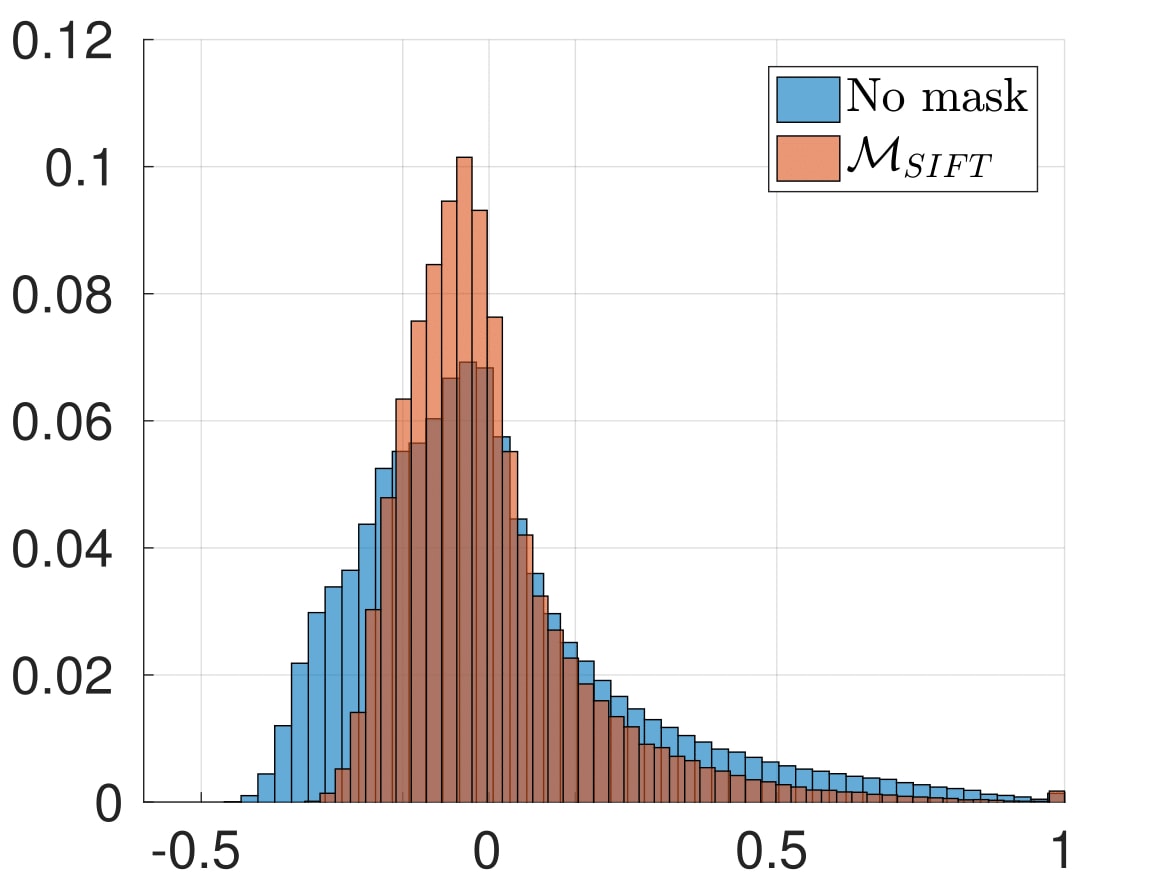}
\caption{}
\label{sfig:sift_hist}
\end{subfigure}
~
\begin{subfigure}[b]{\histscale\textwidth}
\includegraphics[width=\textwidth]{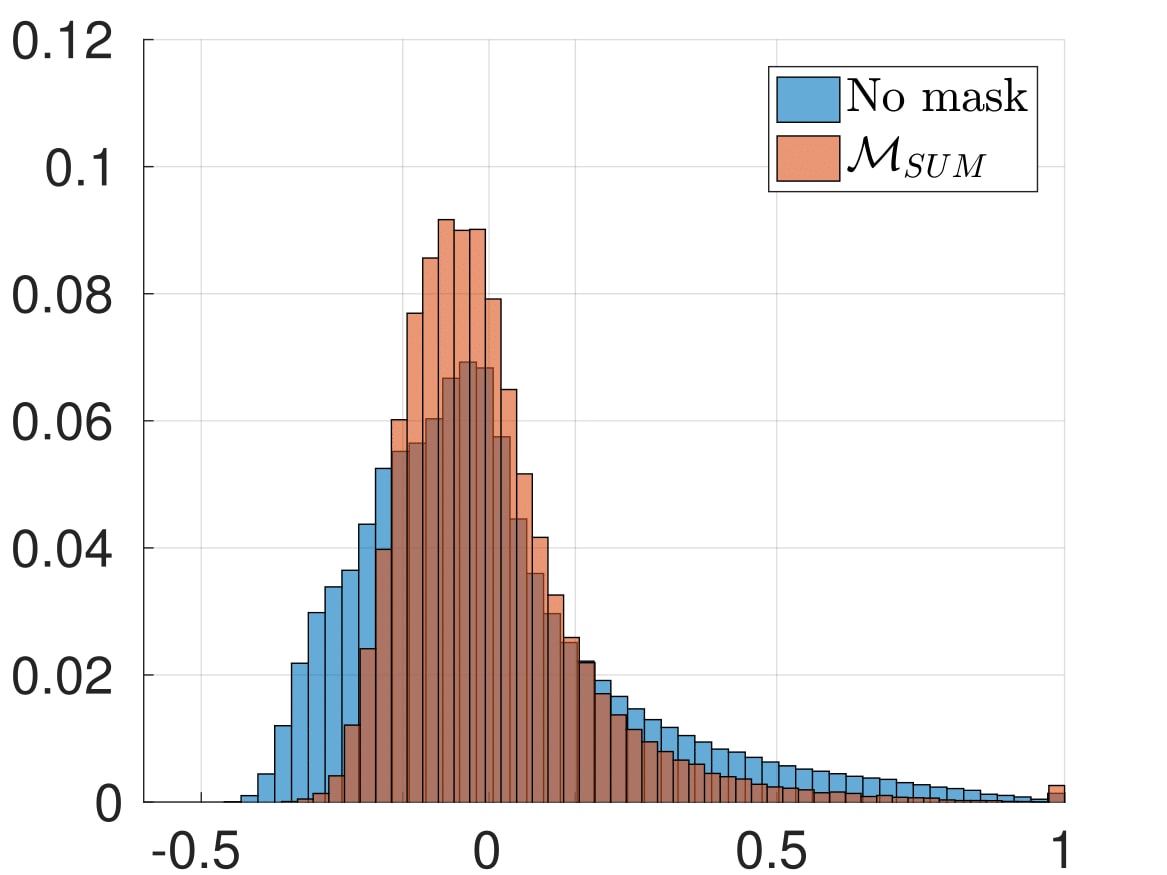}
\caption{}
\label{sfig:sum_hist}
\end{subfigure}
~
\begin{subfigure}[b]{\histscale\textwidth}
\includegraphics[width=\textwidth]{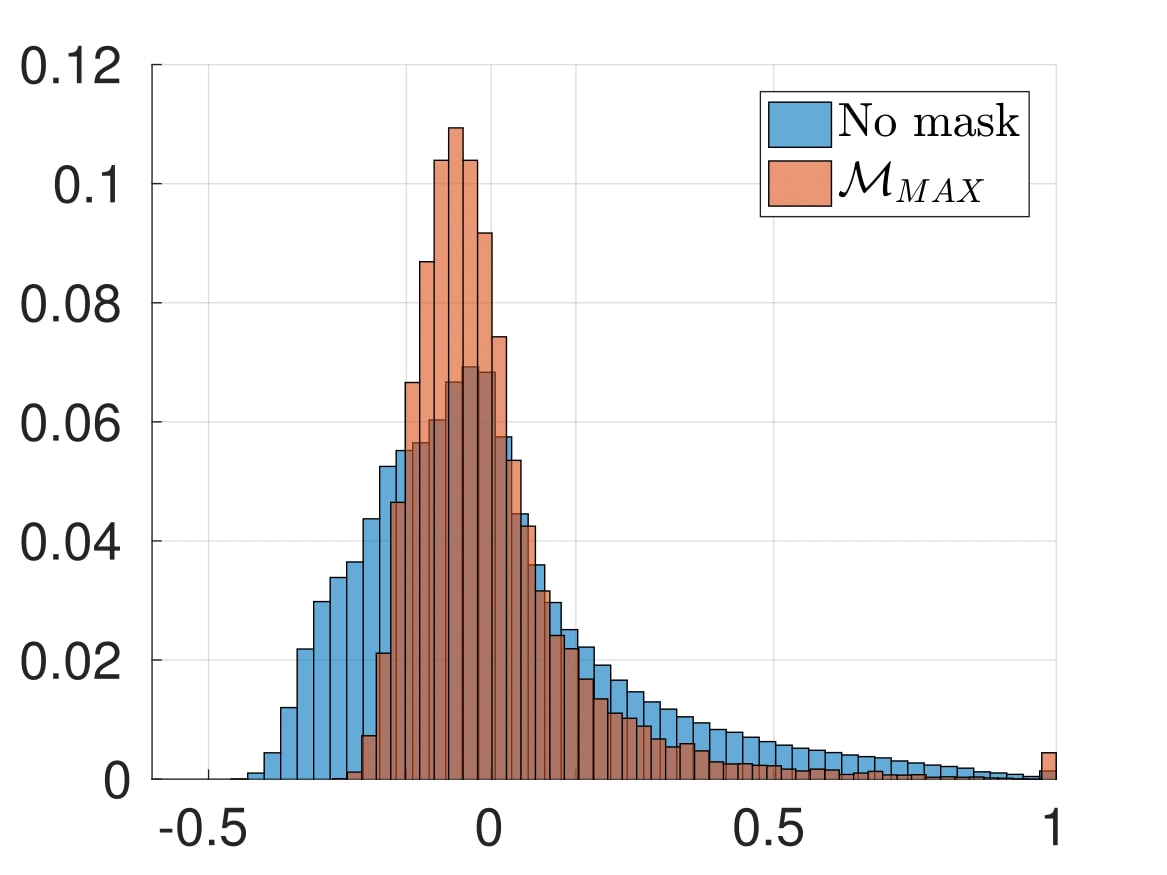}
\caption{}
\label{sfig:max_hist}
\end{subfigure}
~
\begin{subfigure}[b]{\histscale\textwidth}
\includegraphics[width=\textwidth]{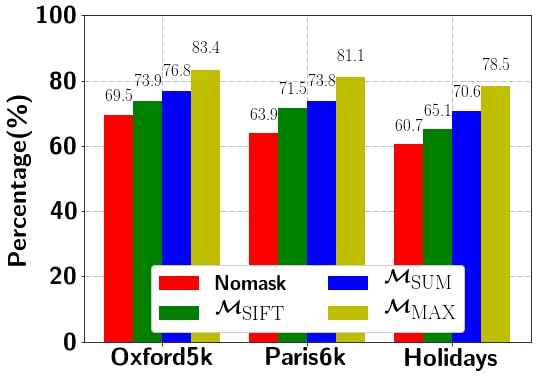}
\caption{}
\label{sfig:std}
\end{subfigure}

\caption{Fig. \ref{sfig:num_local_fea}: The averaged percentage of remaining  local conv. features after applying masks.
Fig. \ref{sfig:sift_hist}, \ref{sfig:sum_hist}, \ref{sfig:max_hist}: Examples of normalized histograms of covariances of sets of local conv. features (from the input image of the first row in Fig. \ref{fig:mask-example}) with/without applying masks. Fig. \ref{sfig:std}: The averaged percentage of the covariance values in the range of $[-0.15, 0.15]$. 
}
\label{fig:hist}
\end{figure*}
\vspace{\reducevspace}

\section{Framework: Embedding and aggregating on selective conv. features}
\label{sec:frame}
\subsection{Pre-processing} 
Given a set $\mathcal{X}=\{\mathbf{x}_{(x,y)}~|~(x,y)\in \mathcal{M}_*\}$, where $\mathcal{M}_* \in \{\mathcal{M}_\text{SUM}, \mathcal{M}_\text{MAX}, $ $ \mathcal{M}_\text{SIFT}\}$ of selective $K$-dimensional local conv. features belonged to the set,
we apply the principal component analysis (PCA) to compress local conv. features to smaller dimension $d$: $\mathbf{x}^{(d)} =M_{\text{PCA}}\mathbf{x}$, where $M_{\text{PCA}}$ is the PCA-matrix. 
There are two reasons for this dimensional reduction operation. Firstly, the lower dimensional local features helps to produce compact final image representation (even applying embedding) as the current trend in image retrieval \cite{cnn_max_pooling,R-MAC,finetune_hard_samples}. 
Secondly, applying PCA could help to remove noise and redundancy; hence, enhancing the discrimination. We subsequently $l2$-normalize the compressed local conv. features. 

\vspace{\reducevspace}
\subsection{Embedding}
\label{ssec:embedding}

In this section, we aim to enhance  the discriminability of selected local conv. features.  
We propose to accomplish this  by embedding the conv. features to higher-dimensional space: $\mathbf{x}\mapsto \phi(\mathbf{x})$, using state-of-the-art embedding methods \cite{FisherVector, vlad, Temb, F-FAemb}. It is worth noting that while in \cite{cnn_max_pooling}, the authors avoid applying embedding on the \textit{original} set of local deep conv. features.  
However, we find that applying the embedding on the set of \textit{selected} features significantly improves their discriminability.  

We brief describle embedding methods used in our work, i.e. Fisher Vector (FV) \cite{FisherVector}, VLAD \cite{vlad}, Temb \cite{Temb}, F-FAemb \cite{F-FAemb}. Note that in the original design of FV and  VLAD, the embedding and the aggregation (i.e., sum aggregation) are integrated. This prevents the using of recent state-of-the-art aggregation (i.e., democratic pooling \cite{Temb}) on the embedded vectors produced by FV, VLAD. Hence, in order to make the embedding and the aggregating flexible, we decompose the formulation of VLAD and FV. Specifically, we apply the embedding on each local feature separately. This allows  different aggregating methods to be applied on the embedded features. 

For clarity, we pre-define the codebook of visual words learning by Gaussian Mixture Model used in Fisher Vector method as $\mathcal{C}_\mathbf{G} = \{\mu_i;\Sigma_i;w_i\}_{i=1}^k$, where $w_i$, $\mu_i$ and $\Sigma_i$ denote respectively the weight, mean vector and covariance matrix of the $i$-th Gaussian. Similarly, the codebook learning by K-means used in VLAD, T-emb, and F-FAemb methods are defined as $\mathcal{C}_\mathbf{K}=\{c_j\}_{j=1}^k$, where $c_j$ is a centroid.

\textbf{\textit{Fisher Vector (FV)}} produces a high-dimensional vector representation of $(2\times k\times d)$-dimension when considering both 1-st and 2-nd order statistic of the local features. 
\begin{equation}
\begin{split}
&\phi_\text{FV}(\mathbf{x}) = \left[\cdots,u_i^T,\cdots, v_i^T,\cdots\right]^T\qquad i=1,\cdots,k
\\ &u_i = \frac{p_{i}(\mathbf{x})}{\sqrt{w_i}}\left( \frac{\mathbf{x} - \mu_{i}}{\sigma_{i}}\right) \qquad  v_i=\frac{p_{i}(\mathbf{x})}{\sqrt{2w_i}}\left[ \left(\frac{\mathbf{x}-\mu_{i}}{\sigma_{i}}\right)^2 - 1 \right]\\
\end{split}
\end{equation}
Where $p_i(\mathbf{x})$ is the posterior probability capturing the strength of relationship between a sample $\mathbf{x}$ and the $i$-th Gaussian model and $\sigma_{i}=\sqrt{\text{diag}(\Sigma_i)}$.

\textbf{\textit{VLAD}} \cite{vlad} 
is considered as a simplification of the FV. It embeds $\mathbf{x}$ to the feature space of $(d\times k)$-dimension.
\begin{equation}
\phi_\text{VLAD}(\mathbf{x}) = [\cdots, q_i(\mathbf{x} - c_i)^T,\cdots]^T\qquad i=1,\cdots, k
\end{equation}
Where $c_i$ is the $i$-th visual word of the codebook $\mathcal{C}_\mathbf{K}$,
$q_i=1$ if $c_i$ is the nearest visual word of $\mathbf{x}$ and $q_i=0$ otherwise.

\textbf{\textit{T-emb}} \cite{Temb}. Different from FV and VLAD, T-emb avoids the dependency on absolute distances by only preserve direction information between a feature $\mathbf{x}$ and visual words $c_i\in \mathcal{C}_\mathbf{K}$.
\begin{equation}
\phi_\Delta(\mathbf{x}) = \left[\cdots,\left(\frac{\mathbf{x}-c_i}{\|\mathbf{x}-c_i\|}\right)^T,\cdots \right]^T\qquad i=1,\cdots, k
\end{equation}

\textbf{\textit{F-FAemb}} \cite{F-FAemb}. Departing from the idea of linearly approximation of non-linear function in high dimensional space, the authors showed that the resulted embedded vector of the approximation process is the generalization of several well-known embedding methods such as VLAD \cite{vlad}, TLCC \cite{TLCC}, VLAT \cite{VLAT}.
\begin{equation}
s_i = \gamma_i(\x) V\left(\left(\x - c_i\right)\left(\x - c_i\right)^T\right) \qquad i=1,\cdots, k
\end{equation}
where $\gamma_i(\x)$ is coefficient corresponding to visual word $c_i$ achieved by the function approximation process and $V(H)$ is a function  that  flattens  the  matrix to  a  vector. The  vectors  $s_i$ are concatenated to form the single embedded feature $\phi_\text{F-FAemb}$. 

\vspace{\reducevspace}
\subsection{Aggregating} 
\label{ssec:aggregating}
Let $\mathcal{X}_\phi = \{\phi(\mathbf{x}_i)\}$ 
be a set of embedded local descriptors. 
Sum/average-pooling and max-pooling are two common methods for aggregating this set to a single global feature of length $D$. 

When using the features generating from the activation function, e.g. ReLU \cite{Alexnet}, of a CNN, \textbf{\textit{sum/average-pooling}} ($\psi_\text{s}/\psi_\text{a}$) lack discriminability because they average the high activated outputs by non-active outputs. Consequently, they weaken the effect of highly activated features.
\textbf{\textit{Max-pooling}} ($\psi_\text{m}$), on the other hand, is more preferable since it only retains the high activation for each visual content.
However, it is worth noting that in practical, the max-pooling is only successfully applied when features are sparse \cite{theoretical_pooling}. For examples, in \cite{R-MAC,finetune_hard_samples}, the max-pooling is applied on each feature map because there are few of high activation values in a feature map.  
When the embedding is applied to embed local features to high dimensional space, the max-pooling may be failed since the local features are no longer sparse  \cite{theoretical_pooling}.

Recently, H. Jegou et. al. \cite{Temb} introduced \textbf{\textit{democratic aggregation}} $(\psi_\text{d})$ method applied to image retrieval problem. 
Democratic aggregation can work out-of-the-box with various embedded features such as VLAD \cite{vlad}, Fisher vector \cite{FisherVector}, T-emb \cite{Temb}, FAemb \cite{FAemb}, F-FAemb \cite{F-FAemb}, and it has been shown to outperform sum-pooling in term of retrieval performance with embedded hand-crafted SIFT features \cite{Temb}. We also conduct experiments for this method on our framework.


\vspace{\reducevspace}
\subsection{Post-processing}
\textbf{\textit{Power-law normalization (PN).}} The \textit{burstiness} of visual elements \cite{burstiness} is known as a major drawback of hand-crafted local descriptors, e.g. SIFT \cite{SIFT_Lowe}, such that numerous descriptors are almost similar within the same image. As a result, this phenomenon strongly affects the measure of similarity between two images. By applying power-law normalization \cite{power-norm} to the final image representation $\psi$ and subsequently $l2$-normalization, it has been shown to be an efficient way to reduce the effect of burstiness \cite{Temb}. The power-law normalization formula is given as $PN(x)=\text{sign}(x)|x^\alpha|$, where $0 \le \alpha \le 1$ is a constant \cite{power-norm}.

\begin{figure}[ht]
\centering
\begin{subfigure}[b]{0.23\textwidth}
\includegraphics[width=\textwidth]{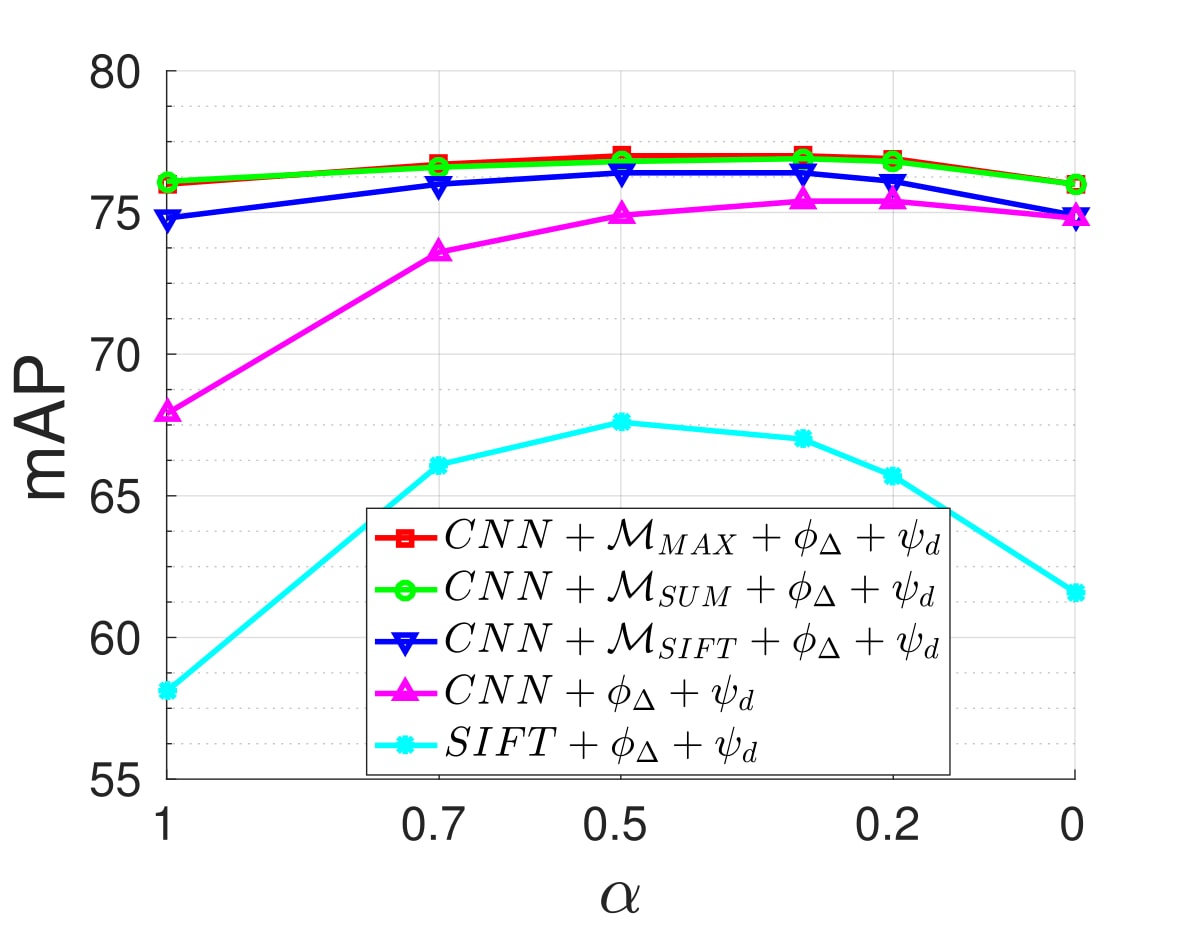}
\caption{\textbf{\textit{Oxford5k}}}
\end{subfigure}
\begin{subfigure}[b]{0.23\textwidth}
\includegraphics[width=\textwidth]{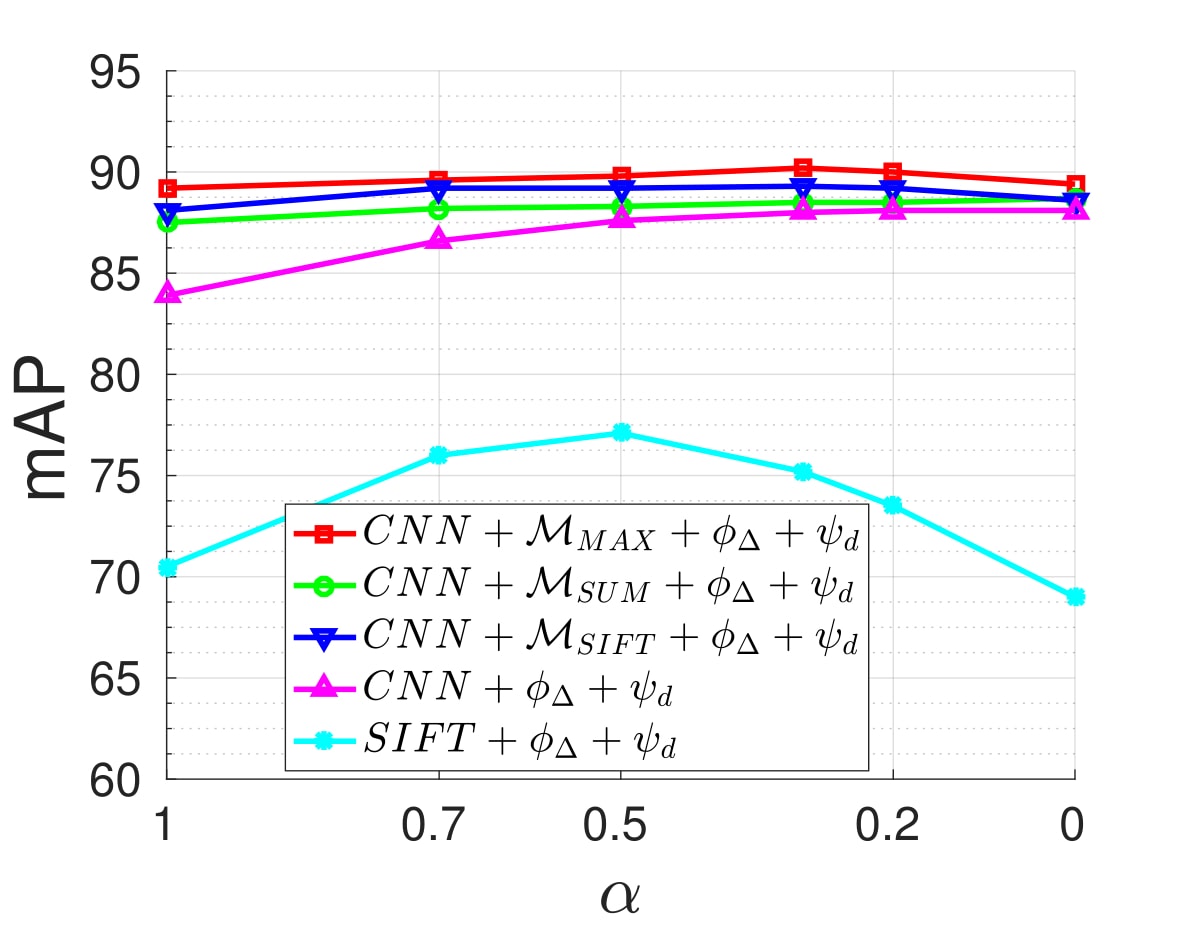}
\caption{\textbf{\textit{Holidays}}}
\end{subfigure}
\caption{Impact of power-law normalization factor $\alpha$ on \textbf{\textit{Oxford5k}} and \textbf{\textit{Holidays}} datasets. 
Following the setting in \cite{Temb}, we set $d=128$ and $|\mathcal{C}|=64$ for both SIFT and conv. features. The local conv. features are extracted from $\mathtt{pool5}$ layer of the pre-trained VGG \cite{VGG}.}
\label{fig:norm-effect}
\end{figure}

However, to the best of our knowledge, no previous work has re-evaluated the \textit{burstiness} phenomena on the local conv. features. 
Figure \ref{fig:norm-effect} shows the analysis of PN effect on local conv. features using various masking schemes. This figure shows that the local conv. features ($CNN+\phi_\Delta+\psi_\text{d} $) are still affected by the burstiness: the retrieval performance changes when applying PN.
The figure also shows that the burstiness has much stronger effect on SIFT features ($SIFT+\phi_\Delta+\psi_\text{d} $) than conv. features. 
The proposed SIFT, SUM and MAX masks help reduce the burstiness effect significantly: the PN has less effect on $CNN+\mathcal{M}_\text{MAX/SUM/SIFT}+ \phi_\Delta+\psi_\text{d}$ than on $CNN+\phi_\Delta+\psi_\text{d}$. This illustrates the capability of removing redundant local features of our proposed masking schemes. Similar to previous work, we set $\alpha=0.5$ in our later experiments.

\vspace{-0.4em}

\textbf{\textit{Rotation normalization and dimension reduction.}} The power-law normalization suppresses visual burstiness but not  frequent co-occurrences which also corrupt the similarity measure \cite{neg_evidence}. In order to reduce the effect of co-occurrences, we follow \cite{neg_evidence,Temb} to rotate data with a whitening matrix learned from the aggregated vectors of the training set. The rotated vectors are used as the final image representations in our image retrieval system. 
\section{Experiments}
\label{sec:exp}
In this section, we will conduct comprehensive experiments to evaluate our proposed framework on three standard image retrieval benchmark datasets, including INRIA Holidays \cite{holiday}, Oxford Buildings dataset \cite{oxford5k}, and Paris dataset \cite{paris6k}.

\subsection{Datasets, Evaluation protocols, and Implementation notes}
The \textbf{INRIA Holidays dataset} (\textit{Holidays}) \cite{holiday} contains 1491 vacation snapshots corresponding to 500 groups of the same instances. The query image set consists of one image from each group. We also manually rotate images (by $\pm 90$ degrees) to fix the wrong image orientation as in \cite{neuralcode, cnn_max_pooling,CroW}. 

The \textbf{Oxford Buildings dataset} (\textit{Oxford5k}) \cite{oxford5k} contains 5063 photographs from Flickr associated with Oxford landmarks. 55 queries corresponding to 11 buildings/landmarks are fixed, and the ground truth relevance of the remaining dataset w.r.t. these 11 classes is provided. Following the standard protocol \cite{R-MAC, deep_img_retr}, we use the cropped query images based on provided bounding boxes.

The \textbf{Paris dataset} (\textit{Paris6k}) \cite{paris6k} are composed of 6412 images of famous landmarks in Paris. Similar to \textit{Oxford5k}, this dataset has 55 queries corresponding to 11 buildings/landmarks. We also use provided bounding boxes to crop the query images accordingly. 

\textbf{Larger datasets.} We additionally use 100k Flickr images~\cite{oxford5k} in combination with \textit{Oxford5k} and \textit{Paris6k} to compose \textit{Oxford105k} and \textit{Paris106k}, respectively. This 100k distractors are to allow  evaluating retrieval methods at larger scale. 

\smallskip
\textbf{Evaluation protocols.} The retrieval performance is reported as mean average precision (\textbf{\textit{mAP}}) over query sets for all datasets. In addition, the \textit{junk} images, which are 
defined as 
unclear to be relevant or not, are removed from the ranking.

\smallskip
\textbf{Implementation notes.} In the image retrieval task, it is important to use held-out datasets to learn all necessary parameters as to avoid overfitting \cite{cnn_max_pooling, finetune_hard_samples,deep_img_retr}. In particular, the set of 5000 Flickr images\footnote{We randomly select 5000 images from the 100.071 Flickr image set \cite{oxford5k}.} is used as the held-out dataset to learn parameters for \textit{Holidays}. Similarly, \textit{Oxford5k} is used for \textit{Paris6k} and \textit{Paris106k}, and \textit{Paris6k} for \textit{Oxford5k} and \textit{Oxford105k}.

All images are resized so that the maximum dimension is 1024 while preserving aspect ratios before fed into the CNN. Additionally, as the common practice in recent works \cite{R-MAC, cnn_max_pooling, finetune_hard_samples, deep_img_retr}, the pretrained VGG16 \cite{VGG} (with Matconvnet \cite{matconvnet} toolbox) is used to extract deep convolutional features. 
We utilize the VLFeat toolbox \cite{vlfeat} for SIFT detector
\footnote{Note that VLFeat toolbox combines both SIFT detector and extractor in a single built-in function.}. 
Additionally, in the rare case of an image with no SIFT feature, the SIFT-mask is ignored and we apply embedding and aggregating for all local features.
We summarize the notations in Table \ref{tb:notation_sum}. Our implementation of the method is available at $\mathtt{https://github.com/hnanhtuan/selectiveConvFeature}$.

\begin{table}[h]
\caption{Notations and their corresponding meanings. $\mathcal{M}, \phi, \psi$ denote masking, pooling and embedding respectively.}
\label{tb:notation_sum}
\centering
\begin{tabular}{|c|l||c|l|}
\hline
Notation & Meaning & Notation & Meaning \\
\hline
$\mathcal{M}_\text{SIFT}$ & SIFT-mask & $\psi_\text{a}$ & Average-pooling\\
$\mathcal{M}_\text{SUM}$ & SUM-mask & $\psi_\text{s}$ & Sum-pooling\\
$\mathcal{M}_\text{MAX}$ & MAX-mask & $\psi_\text{d}$ & Democratic-pooling \cite{Temb}\\
$\phi_\text{FV}$ & FV \cite{FisherVector} & $\phi_\text{VLAD}$ & VLAD \cite{vlad} \\
$\phi_\Delta$ & T-emb \cite{Temb} & $\phi_\text{F-FAemb}$ & F-FAemb \cite{F-FAemb} \\
\hline
\end{tabular}
\end{table}

\subsection{Effects of parameters}
\subsubsection{Framework.} 
In this section, we will conduct experiment to comprehensively compare various embedding and aggregating frameworks in combination with different proposed masking schemes. To make a fair comparison, we empirically set the retained PCA components-$d$ and size of the visual codebooks-$|\mathcal{C}|$ so as to produce the same final feature dimensionality-$D$ as mentioned in Table \ref{tb:emb-setting}. Note that, as proposed in original papers, F-FAemb \cite{F-FAemb} method requires to remove first $d(d+1)/2$ components of the features after aggregating step (Section \ref{ssec:aggregating}). However, we empirically found that truncating the first $d(d+1)$ components generally achieves better performances.

\begin{table}[ht]
\caption{Configuration of different embedding methods.}
\label{tb:emb-setting}
\centering
\begin{tabular}{|l|c|c|c|}
\hline
\textbf{Method} & \textbf{PCA-}$d$ & $|\mathcal{C}|$ & $D$ \\
\hline
FV \cite{FisherVector} & 48 & 44 & $2\times d\times |\mathcal{C}|=4224$\\
VLAD \cite{vlad} & 64 & 66 & $d\times |\mathcal{C}|=4224$  \\
T-emb \cite{Temb} & 64 & 68 & $d\times |\mathcal{C}|-128=4224$ \\
F-FAemb \cite{F-FAemb} & 32 & 10 & $\displaystyle{\frac{(|\mathcal{C}|-2)\times d\times (d+1)}{2}=4224}$\\
\hline
\end{tabular}
\end{table}

The comparison results on \textit{Oxford5k}, \textit{Paris6k}, and \textit{Holidays} datasets are reported in Table \ref{tb:compare_pooling_framework}. Firstly, we can observe that the democratic pooling \cite{Temb} clearly outperforms sum/average-pooling on both FV \cite{FisherVector} and VLAD \cite{vlad} embedding methods. Secondly, our proposed masking schemes help to achieve considerable gains in performance across the variety of embedding and aggregating frameworks. Additionally, the MAX-mask generally provides the higher performance boosts than the SUM/SIFT-mask, while SUM-mask and SIFT-mask give comparable results. At the comparison dimensionality-$D=4224$, the framework of $\phi_\Delta+\psi_\text{d}$ and $\phi_\text{F-FAemb}+\psi_\text{d}$ achieves comparable performances across various masking schemes and datasets. In this paper, since $\mathcal{M}_\text{MAX}+ \phi_\Delta+\psi_\text{d}$ provides the best performance, slightly better than $\mathcal{M}_\text{MAX}+ \phi_\text{F-FAemb}+\psi_\text{d}$, we choose $\mathcal{M}_* + \phi_\Delta+\psi_\text{d}$ as our default framework.

\begin{table}[ht]
\caption{Comparison of different frameworks. The ``Bold'' values indicates the best performance in each masking method and the \underline{``Underline''} values indicates best performance across all settings.}
\label{tb:compare_pooling_framework}
\centering
\begin{tabular}{|c|l|c|c|c|c|}
\hline
& \textbf{Method} & $\mathcal{M}_\text{MAX}$ & $\mathcal{M}_\text{SUM}$ & $\mathcal{M}_\text{SIFT}$ & None \\
\hline
\multirow{6}{*}{\rotatebox[origin=c]{90}{\textbf{Oxford5k}}} & $\phi_\text{FV}+\psi_\text{a}$ & 67.8 & 65.1 &  65.5 & 59.5 \\
& $\phi_\text{FV}+\psi_\text{d}$ & 72.2 & 71.8 & 72.0 & 69.6 \\
& $\phi_\text{VLAD}+\psi_\text{s}$ & 66.3 & 65.6 & 66.4 & 65.1 \\
& $\phi_\text{VLAD}+\psi_\text{d}$ & 69.2 & 70.5 & 71.3 & 69.4 \\
& $\phi_\Delta+\psi_\text{d}$ & \underline{\textbf{75.8}} & \textbf{75.7} & \textbf{75.3} & 73.4 \\
& $\phi_\text{F-FAemb}+\psi_\text{d}$ & 75.2 & 74.7 & 74.4 & \textbf{73.8} \\
\hline
\multirow{6}{*}{\rotatebox[origin=c]{90}{\textbf{Paris6k }}} & $\phi_\text{FV}+\psi_\text{a}$ & 78.4 & 76.4 & 75.8 & 68.0 \\
& $\phi_\text{FV}+\psi_\text{d}$ & 84.5 & 82.2 & 82.4 & 76.9 \\
& $\phi_\text{VLAD}+\psi_\text{s}$ & 77.7 & 74.5 & 76.0 & 73.2 \\
& $\phi_\text{VLAD}+\psi_\text{d}$ & 80.3 & 79.5 & 81.3 & 79.3 \\
& $\phi_\Delta+\psi_\text{d}$ & \underline{\textbf{86.9}} & 84.8 & 85.3 & \textbf{83.9} \\
& $\phi_\text{F-FAemb}+\psi_\text{d}$ & 86.6 & \textbf{85.9} & \textbf{85.6} & 82.9 \\
\hline
\multirow{6}{*}{\rotatebox[origin=c]{90}{\textbf{Holidays }}} & $\phi_\text{FV}+\psi_\text{a}$ & 83.2 & 80.0 & 81.5 & 78.2 \\
& $\phi_\text{FV}+\psi_\text{d}$ & 87.8 & 86.7 & 87.1 & 85.2 \\
& $\phi_\text{VLAD}+\psi_\text{s}$ & 83.3 & 82.0 &  83.6 & 82.7 \\
& $\phi_\text{VLAD}+\psi_\text{d}$ & 85.5 & 86.4 & 87.5 & 86.1 \\
& $\phi_\Delta+\psi_\text{d}$ & \underline{\textbf{89.1}} & 88.1 & \textbf{88.6} & \textbf{87.3} \\
& $\phi_\text{F-FAemb}+\psi_\text{d}$ & 88.6 & \textbf{88.4} & 88.5 & 86.4\\
\hline
\end{tabular}
\end{table}
\vspace{-0.4em}
\subsubsection{Final feature dimensionality} 
\label{sssec:final-dim}
Different from recent works using convolutional features \cite{cnn_max_pooling,CroW,finetune_hard_samples,R-MAC}, which have the final feature dimensionality upper bounded by the number of output feature channel $K$ of network architecture and selected layer, e.g. $K=512$ for $\mathtt{Conv5}$ of VGG \cite{VGG}. Taking the advantages of embedding methods, similar to NetVLAD \cite{netvlad}, our proposed framework provides more flexibility on choosing the length of final representation.

Considering our default framework - $\mathcal{M}_* + \phi_\Delta+\psi_\text{d}$, we empirically set the number of retained PCA components and the codebook size when varying the dimensionality in Table \ref{tb:vary-dim}. For compact final representations, we choose $d=32$ to avoid using too few visual words since this drastically degrades performance \cite{Temb}. For longer final representations, imitating the setting for SIFT in \cite{Temb}, we reduce local conv. features to $d=128$ and set $|\mathcal{C}|=64$. Note that the settings in Table \ref{tb:vary-dim} are applied for all later experiments.
\begin{table}[ht]
\caption{Number of retained PCA components and codebook size when varying the dimensionality.}
\label{tb:vary-dim}
\centering
\begin{tabular}{|l|c|c|c|c|c|}
\hline
\textbf{Dim.} $D$ & 512-D & 1024-D & 2048-D & 4096-D & 8064-D\\
\hline
\textbf{PCA} $d$ & 32 & 64 & 64 & 64 & 128 \\
$|\mathcal{C}|$ & 20 & 18 & 34 & 66 & 64 \\
\hline
\end{tabular}
\end{table}

\begin{figure}[ht]
\centering
\begin{subfigure}[b]{0.23\textwidth}
\includegraphics[width=\textwidth]{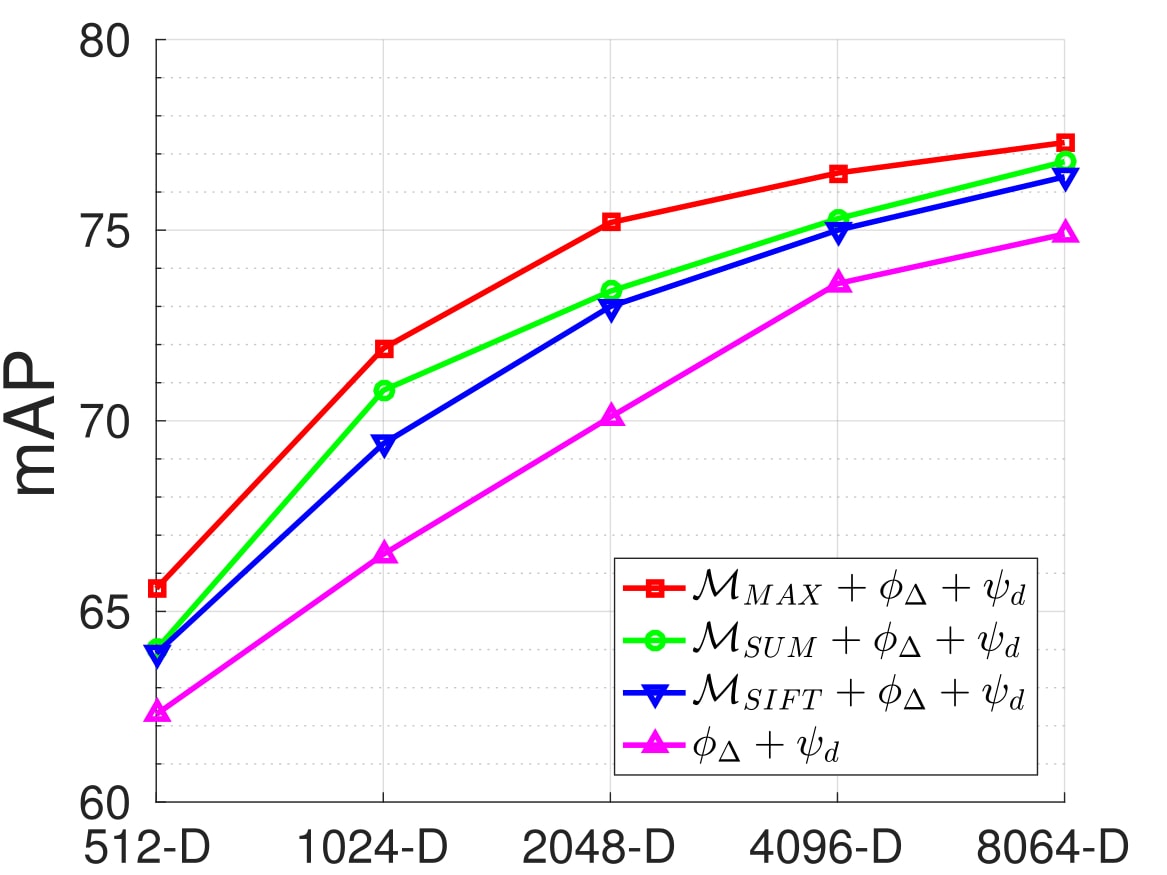}
\caption{\textbf{\textit{Oxford5k}}}
\end{subfigure}
\begin{subfigure}[b]{0.23\textwidth}
\includegraphics[width=\textwidth]{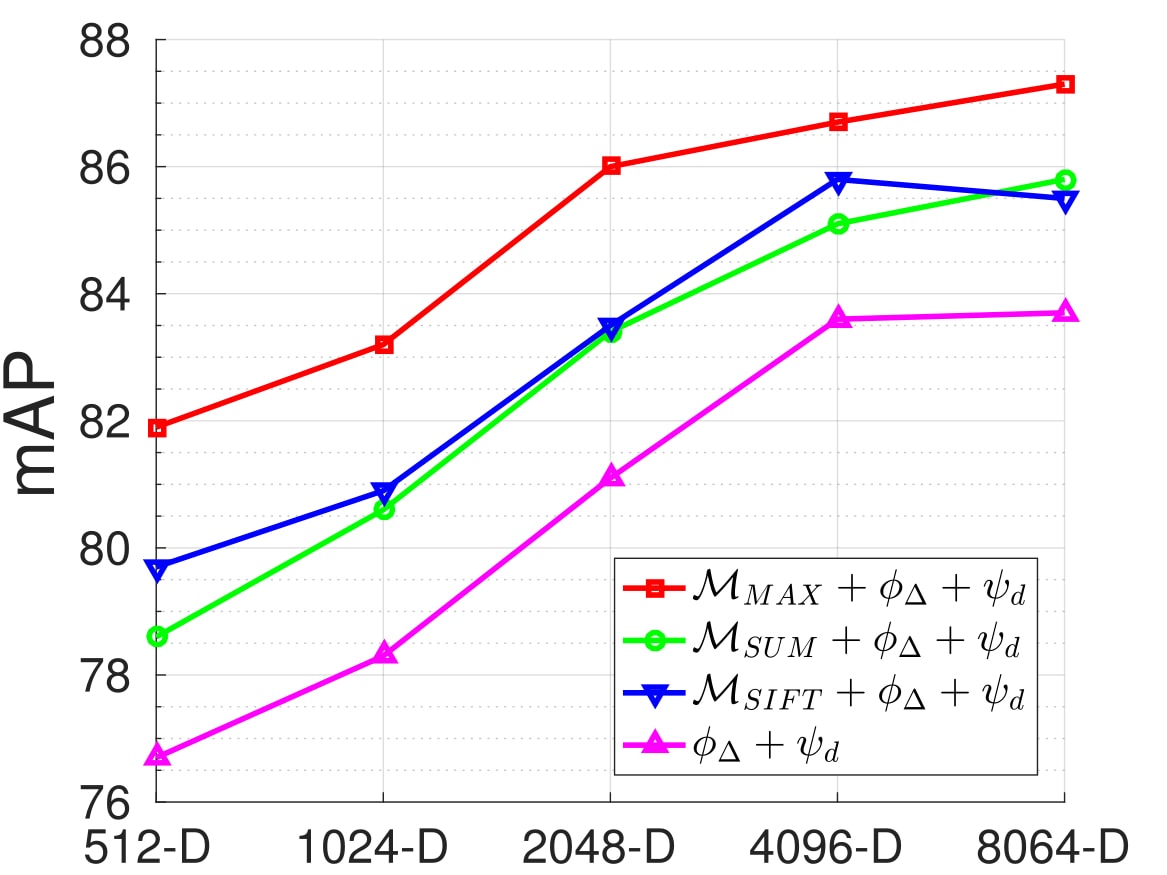}
\caption{\textbf{\textit{Paris6k}}}
\end{subfigure}
\caption{Impact of the final representation dimensionality on \textbf{\textit{Oxford5k}} and \textbf{\textit{Paris6k}} datasets.}
\label{fig:compare-mask-vary-D}
\end{figure}

The Figure \ref{fig:compare-mask-vary-D} shows the retrieval performance of two datasets, {\textit{Oxford5k}} and {\textit{Paris6k}}, when varying the final feature dimensionality. Obviously, our proposed method can significantly boost the performance when increasing the final feature dimensionality. In addition, we also observe that the masking schemes consistently help to gain extra performance across different dimensionalities.

\vspace{\reducevspace}
\subsubsection{Image size}
Even though the authors in \cite{R-MAC,CroW} found that the original size of images ($\max(W_I,H_I)=1024$) provides higher performance, it is important to evaluate our method with a smaller image size on the performance since our method depends on the number of local conv. features. Table \ref{tb:compare-image-size} shows the retrieval performance of \textit{Oxford5k} and \textit{Paris6k} datasets with the image size of $\max(W_I,H_I)=1024$ and $\max(W_I,H_I)=724$. Similar to the reported results of \cite{R-MAC} on {\textit{Oxford5k}} dataset, we observe around 6-7\% drop in \textbf{\textit{mAP}} when scaling down images to $\max(W_I,H_I)=724$ rather than the original images. While on {\textit{Paris6k}} dataset, interestingly, the performances are more stable to the image size. We also observe a small drop of 2.2\% on \textit{Paris6k} dataset for R-MAC \cite{R-MAC} with our implementation. These suggest that our method and R-MAC method \cite{R-MAC} equivalently affected by the change in the image size.

The performance drops on {\textit{Oxford5k}} can be explained that with bigger images, the CNN can take a closer \textit{``look''} on smaller details in the images. Hence, the local conv. features can better distinguish details in different images. 
While the stable on \textit{Paris6k} dataset can be perceived that the differences on these scenes are at global structures rather than small details as on \textit{Oxford5k} dataset. 


\begin{table}[h]
\caption{Impact of input image size on \textbf{\textit{Oxford5k}} and \textbf{\textit{Paris6k}} datasets. The framework of $\mathcal{M}_\text{MAX/SUM} + \phi_\Delta + \psi_\text{d}$ is used to produce image representations.}
\label{tb:compare-image-size}
\centering
\begin{tabular}{|c|c|c|c|c|c|}
\hline
\multirow{2}{*}{\textbf{Dim.} $D$} & \multirow{2}{*}{$\max(W_I,H_I)$} & \multicolumn{2}{c|}{\textbf{Oxford5k}} &  \multicolumn{2}{c|}{\textbf{Paris6k}} \\
\cline{3-6} & & $\mathcal{M}_\text{SUM}$ & $\mathcal{M}_\text{MAX}$ & $\mathcal{M}_\text{SUM}$ & $\mathcal{M}_\text{MAX}$ \\
\hline
\multirow{2}{*}{512} & 724 & 56.4 & 60.9 & 79.3 & 81.2\\
                     & 1024 & 64.0 & 65.7 & 78.6 & 81.6  \\
\hline
\end{tabular}
\end{table}
\vspace{\reducevspace}
\subsubsection{Layer selection} 
\begin{figure}[ht]
\centering
\begin{subfigure}[b]{0.21\textwidth}
\includegraphics[width=\textwidth]{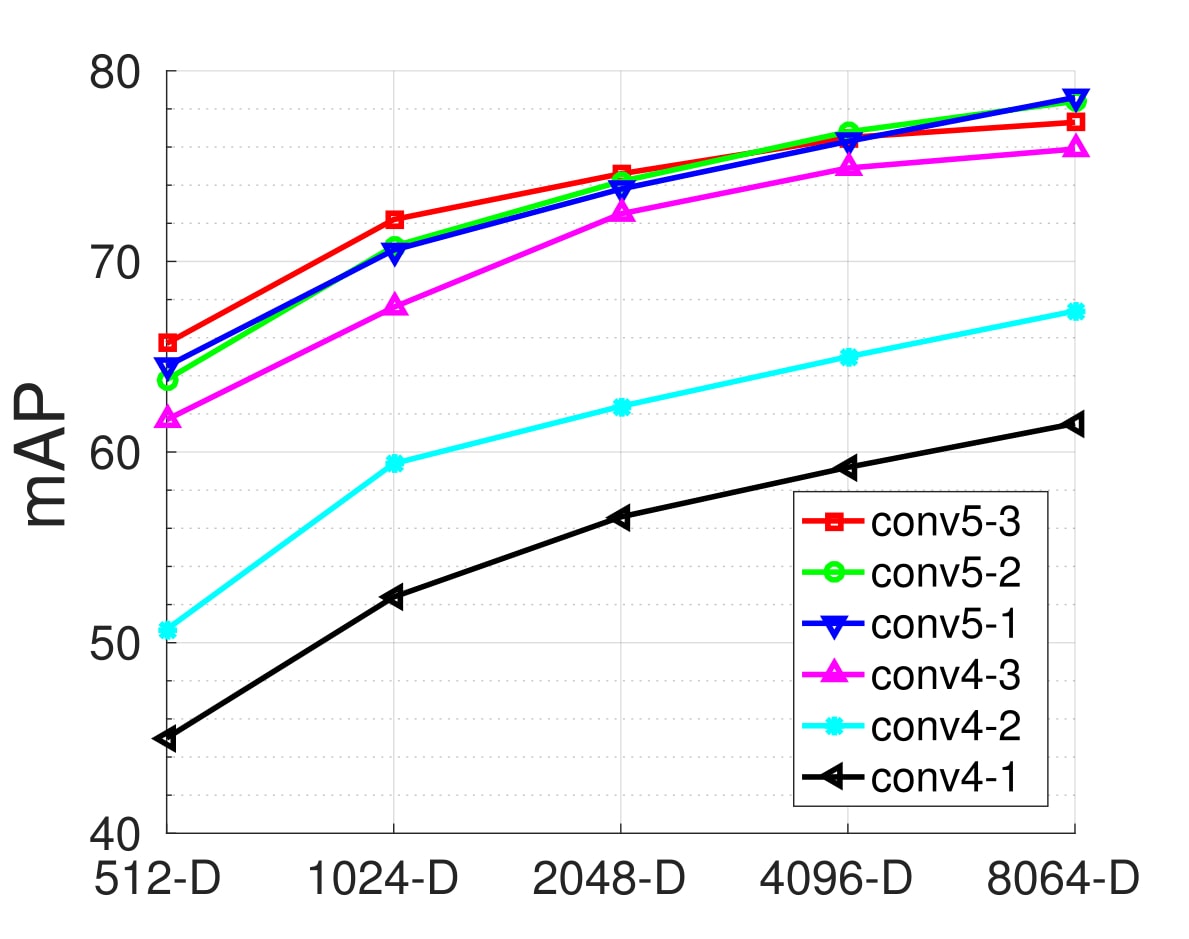}
\caption{\textbf{\textit{Oxford5k}}}
\end{subfigure}
\begin{subfigure}[b]{0.21\textwidth}
\includegraphics[width=\textwidth]{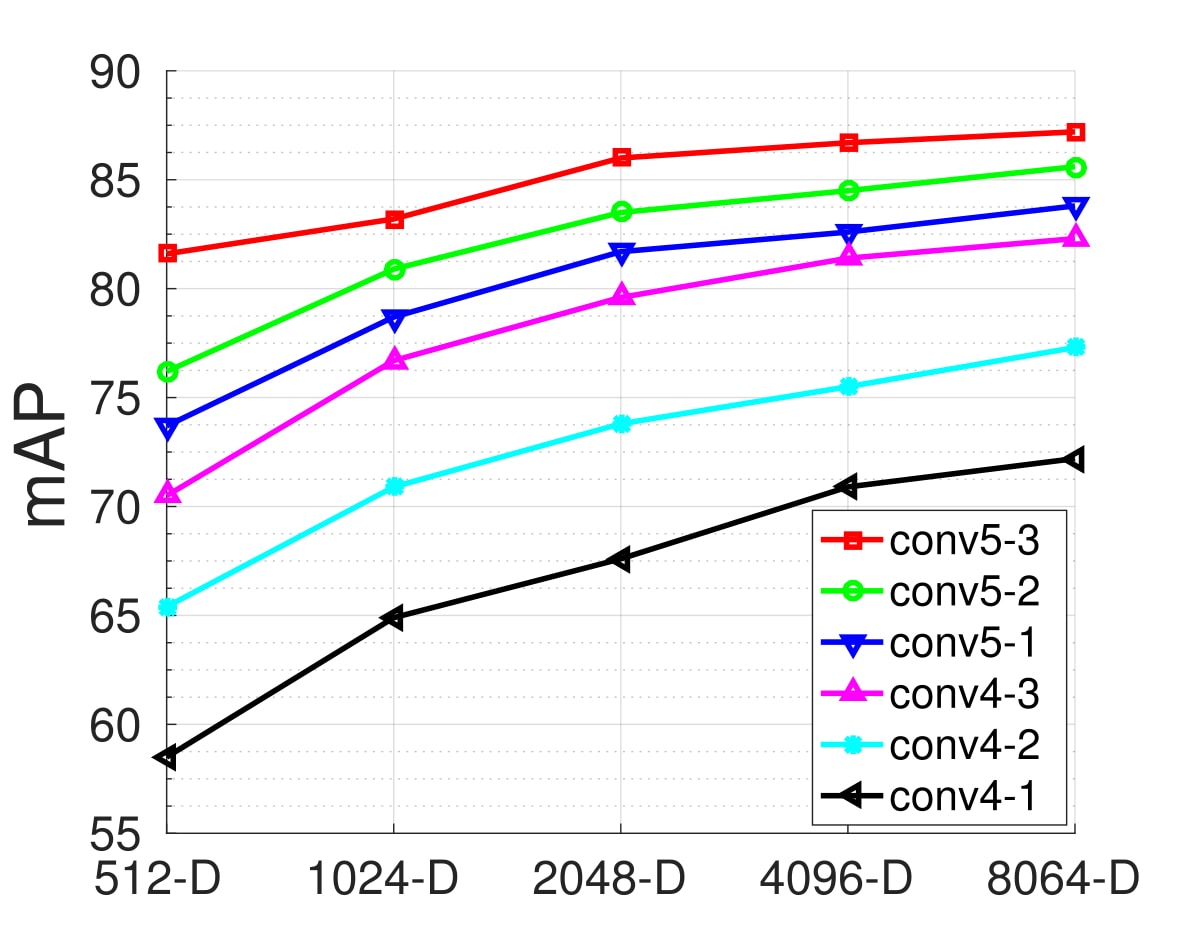}
\caption{\textbf{\textit{Paris6k}}}
\end{subfigure}
\caption{Evaluation of retrieval performance of local deep conv. features from different layers on \textbf{\textit{Oxford5k}} and \textbf{\textit{Paris6k}} datasets. The framework of $\mathcal{M}_\text{MAX} + \phi_\Delta + \psi_\text{d}$ is used to produce image representations.}
\label{fig:compare-layers}
\end{figure}

In \cite{cnn_max_pooling}, while evaluating at different feature lengths, the authors claimed that deeper conv. layer produces features with more reliable similarities. Hence, we want to re-evaluate this statement by comparing the retrieval performance (\textbf{\textit{mAP}}) of features extracted from different conv. layers at the same dimensionality. In this experiment, we extract features from different conv. layers, including $\mathtt{conv5\text{-}3}$, $\mathtt{conv5\text{-}2}$, $\mathtt{conv5\text{-}1}$, $\mathtt{conv4\text{-}3}$, $\mathtt{conv4\text{-}2}$, and $\mathtt{conv4\text{-}1}$, following by a $2\times 2 ~\mathtt{max}\text{-}\mathtt{pool}$ layer with stride of 2. 
The results of our comprehensive experiments on \textit{Oxford5k} and \textit{Paris6k} datasets are presented in  Figure \ref{fig:compare-layers}. We can observe that there are small drops in performance when using lower conv. layers until $\mathtt{conv4\text{-}3}$. When going down further to $\mathtt{conv4\text{-}2}$ and $\mathtt{conv4\text{-}1}$, there are significant drops in performance.
Regarding the pre-trained VGG network \cite{VGG}, this fact indicates that the last conv. layer produces the most reliable representation for image retrieval.

\vspace{\reducevspace}
\begin{table*}[!t]
\caption{Comparison with the state of the art.}
\label{tb:state-of-art}
\centering
\begin{tabular}{|c|l|c|c|c|c|c|c|}
\hline
 & \multirow{2}{*}{\textbf{Method}} & \multirow{2}{*}{\textbf{Dim.}} & \multicolumn{5}{c|}{\textbf{Datasets}}\\
\cline{4-8} & & & \textit{\textbf{Oxford5k}} & \textbf{\textit{Oxford105k}} & \textbf{\textit{Paris6k }}& \textbf{\textit{Paris106k}} & \textbf{\textit{Holidays}} \\
\hline
\hline
\multirow{14}{*}{\rotatebox[origin=c]{90}{\textbf{Off-the-shell network}}} & SPoC \cite{cnn_max_pooling} &  256 & 53.1 & - & 50.1 & - & 80.2 \\
& MOP-CNN \cite{MOP} & 512 & - & - & - & - & 78.4 \\
& CroW \cite{CroW} & 512 & 70.8 & 65.3 & 79.7 & 72.2 & 85.1 \\
& MAC \cite{finetune_hard_samples} & 512 & 56.4 & 47.8 & 72.3 & 58.0 & 76.7\\
& R-MAC \cite{R-MAC} & 512 & 66.9 & 61.6 & 83.0 & 75.7 & - \\
& NetVLAD \cite{netvlad} & 1024 & 62.6 & - & 73.3 & - & 87.3 \\
\cline{2-8}
& \textbf{$\mathcal{M}_\text{SIFT} + \phi_\Delta+\psi_\text{d}$} & 512 & 64.4 & 59.4 & 79.5 & 70.6 & 86.5 \\
& \textbf{$\mathcal{M}_\text{SUM} + \phi_\Delta+\psi_\text{d}$} & 512 & 64.0 & 58.8 & 78.6 & 70.4 & 86.4 \\
& \textbf{$\mathcal{M}_\text{MAX} + \phi_\Delta+\psi_\text{d}$} & 512 & 65.7 & 60.5 & 81.6 & 72.4 & 85.0 \\

& \textbf{$\mathcal{M}_\text{SIFT} + \phi_\Delta+\psi_\text{d}$} & 1024 & 69.9 & 64.3 & 81.7 & 73.8 & 87.1 \\
& \textbf{$\mathcal{M}_\text{SUM} + \phi_\Delta+\psi_\text{d}$} & 1024 & 70.8 & 64.4 & 80.6 & 73.8 & 86.9 \\
& \textbf{$\mathcal{M}_\text{MAX} + \phi_\Delta+\psi_\text{d}$} & 1024 & 72.2 & 67.9 & 83.2 & 76.1 & 88.4\\
\cline{2-8}
& NetVLAD \cite{netvlad} & 4096 & 66.6 & - & 77.4 & - & 88.3 \\
& \textbf{$\mathcal{M}_\text{MAX} + \phi_\Delta+\psi_\text{d}$} & 4096 & 75.3 & 71.4 & 86.7 & 80.6 & 89.0\\
\hline
\hline

\multirow{8}{*}{\rotatebox[origin=c]{90}{\textbf{Finetuned network}}} & siaMAC + R-MAC \cite{finetune_hard_samples} & 512 & 77.0 & 69.2 & 83.8 & 76.4 & 82.5 \\
& NetVLAD fine-tuned \cite{netvlad} & 1024 & 69.2 & - & 76.5 & - & 86.5 \\
\cline{2-8}
& siaMAC $\dagger$ \cite{finetune_hard_samples} + \textbf{$\mathcal{M}_\text{MAX} + \phi_\Delta+\psi_\text{d}$} & 512 & 77.7 & 72.7 & 83.2 & 76.5 & 86.3 \\
& siaMAC $\dagger$ \cite{finetune_hard_samples} + \textbf{$\mathcal{M}_\text{MAX} + \phi_\Delta+\psi_\text{d}$} & 1024 & 81.4 & 77.4 & 84.8 & 78.9 & 88.9 \\
& NetVLAD $\star$ \cite{netvlad} + \textbf{$\mathcal{M}_\text{MAX} + \phi_\Delta+\psi_\text{d}$} & 1024 & 75.2 & 71.7 & 84.4 & 76.9 & 91.5 \\
\cline{2-8}
&  NetVLAD fine-tuned \cite{netvlad} & 4096 & 71.6 & - & 79.7 & - & 87.5 \\
& NetVLAD $\star$ \cite{netvlad} +  \textbf{$\mathcal{M}_\text{MAX} + \phi_\Delta+\psi_\text{d}$} & 4096 & 78.2 & 75.7 & 87.8 & 81.8 & 92.2 \\
& siaMAC $\dagger$ \cite{finetune_hard_samples} + \textbf{$\mathcal{M}_\text{MAX} + \phi_\Delta+\psi_\text{d}$} & 4096 & 83.8 & 80.6 & 88.3 & 83.1 & 90.1 \\
\hline

\end{tabular}
\end{table*}
\subsection{Comparison to the state of the art}

We thoroughly compare our proposed framework with state-of-art methods in image retrieval task. We report experimental results in Table \ref{tb:state-of-art}. 

\textbf{Using off-the-shelf VGG network \cite{VGG}.} At dimensionality of 1024, our method using MAX-mask ($\mathcal{M}_\text{MAX} + \phi_\Delta+\psi_\text{d}$) achieves the highest \textbf{\textit{mAP}} of all compared methods \cite{cnn_max_pooling,CroW,finetune_hard_samples,R-MAC,netvlad} with pre-trained VGG16 network \cite{VGG} across different datasets. Note that some compared methods, e.g. \cite{cnn_max_pooling,CroW,finetune_hard_samples,R-MAC}, have the dimensionality of 512 or 256. This is because the final feature dimensionality of these methods is upper bounded by the number of output feature channel $K$ of network architecture and selected layer, e.g. $K=512$ for $\mathtt{Conv5}$ of VGG16. While our proposed method provides more flexibility in the feature dimensional length.
Furthermore, as discussed in Section \ref{sssec:final-dim}, when increasing the final representation length, our methods can gain extra performance. In particular, at the dimensionality of 4096, our method is very competitive with  methods that require complicated data collection process and days of re-training on powerful GPU \cite{finetune_hard_samples,netvlad}. Our results at 4096-D are lower than \cite{finetune_hard_samples} in \textit{Oxford5k} while higher by a fair margin in \textit{Paris6k} and \textit{Holidays}.

Note that it is unclear in the performance gain when increasing the length of the final representation in R-MAC \cite{R-MAC} or CRoW \cite{CroW}, even at the cost of a significant increase in the number of CNN parameters and the additional efforts of re-training. 
In fact, in \cite{generic2specific}, the authors design an experiment to investigate whether increasing the number of conv. layers, before the fully connected one from which the representation is extracted, would help increase the performance of various visual tasks, including image classification, attribute detection, fine-grained recognition, compositional, and instance retrieval. Interestingly, the experimental results show that while the performance increases on other tasks, it degrades on the retrieval one. The authors explain that the more powerful the network is, the more generality it can provide. As a result, the representation becomes more invariant to instance level differences. Even though, in this experiment, the image representation is constructed from a fully-connected layer, which is different from our current context using conv. layer, the explanation in \cite{generic2specific} could still be applicable. This raises the question about the efficiency of increasing number of channels in a conv. layer as a way to increase final representation dimensionality in SPoC \cite{cnn_max_pooling}, R-MAC \cite{R-MAC}, or CRoW \cite{CroW}. 

Regarding NetVLAD \cite{netvlad} and MOP-CNN \cite{MOP}, these methods also can produce higher-dimensional representation. However, at a certain length, our method clearly achieves higher retrieval performance.

\textbf{Taking advantages of fine-tuned VGG network.} Since our proposed methods take the 3D activation tensor of a conv. layer as the input, our framework is totally compatible with fine-tuned networks \cite{finetune_hard_samples,netvlad}.  In the \textbf{Fine-tuned network} section of Table \ref{tb:state-of-art}, we evaluate our best framework - $\mathcal{M}_\text{MAX} + \phi_\Delta+\psi_\text{d}$ - with the local conv. features of fine-tuned VGG for image retrieval task from \cite{netvlad,finetune_hard_samples} as input. ``NetVLAD $\star$'' and ``siaMAC $\dagger$'' mean that the fine-tuned VGG from NetVLAD \cite{netvlad} and siaMAC \cite{finetune_hard_samples} respectively are used to extracted local conv. features. Additionally, ``NetVLAD fine-tuned'' represents the results reported in \cite{netvlad} after fine-tuning for differentiating the results using the off-the-shelf VGG network.

When using local conv. features extracted from fine-tuned network from \cite{finetune_hard_samples}, our method can achieve very competitive results with those from \cite{finetune_hard_samples} at dimensionality of 512. Our method outperforms \cite{finetune_hard_samples} in majority of benchmark datasets, including {\textit{Oxford5k}}, {\textit{Oxford105k}}, {\textit{Holidays}}, and {\textit{Paris106k}}. Furthermore, at 1024 dimensionality, our method outperforms the most competitive method \cite{netvlad,finetune_hard_samples} by more than $+2.5\%$, except {\textit{Paris6k}} dataset with $+1.0\%$ performance gain, to the next best \textbf{\textit{mAP}} values. It is important to note that the end-to-end training architecture proposed in \cite{finetune_hard_samples} still inherits the drawback of upper-bounded final representation dimensionality from R-MAC \cite{R-MAC}.


\subsection{Processing time}
\label{ssec:processing_time}
\begin{figure}[ht]
\centering
\begin{subfigure}[b]{0.20\textwidth}
\includegraphics[width=\textwidth]{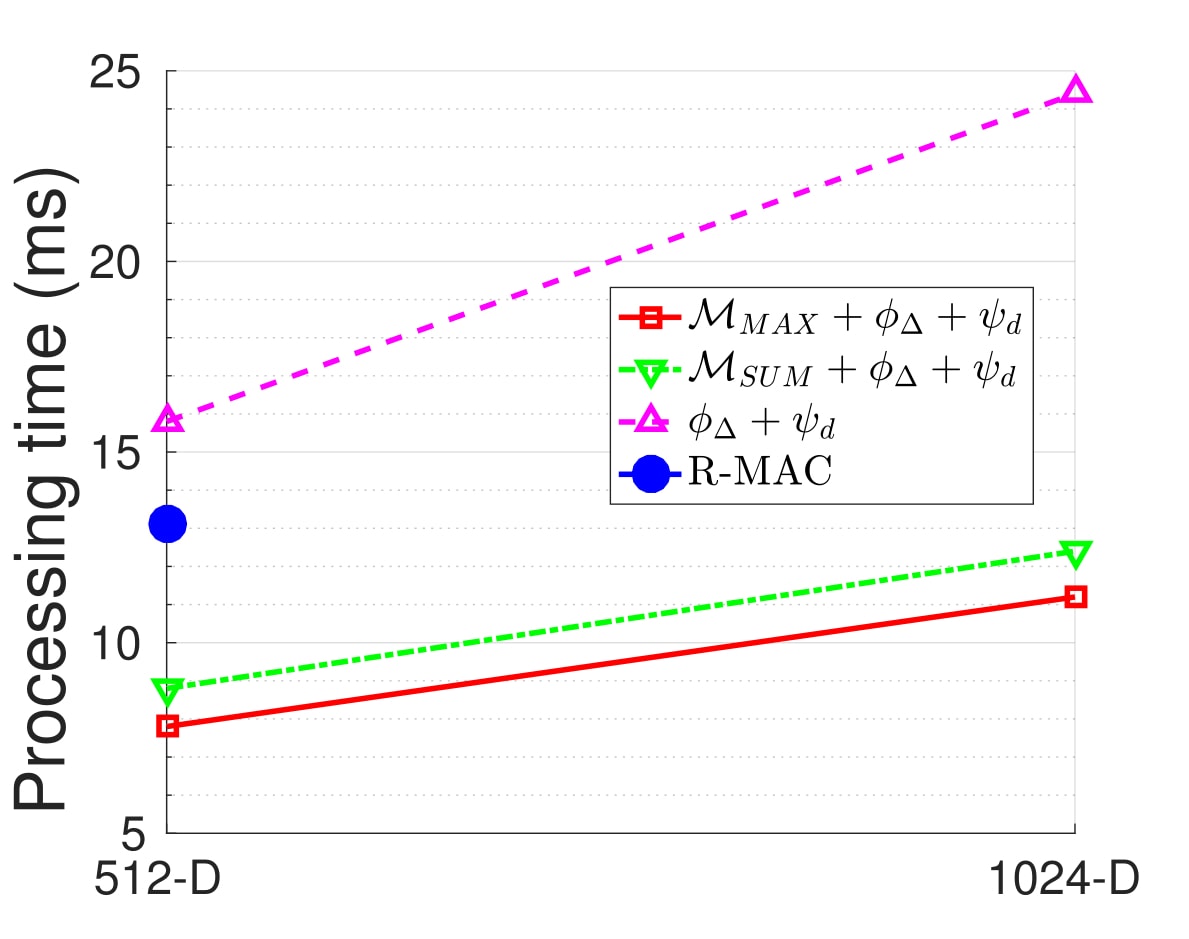}
\end{subfigure}
\caption{The averaged online processing time of 5063 images of \textit{Oxford5k} dataset.}
\label{fig:processing-time}
\end{figure}

We empirically evaluate the online processing time of our proposed framework. We also compare the online processing time between our proposed framework and one of the most competitive methods\footnote{We do not evaluate online processing time for CRoW \cite{CroW} as its published codes are in Python, and it is not appropriate to directly  compare with the Matlab implementation of our method.}: 
R-MAC \cite{R-MAC}. The experiments are carried out on a processor core (i7-6700 CPU @ 3.40GHz). The reported processing time in Figure \ref{fig:processing-time} is the averaged online processing times of $5063$ images of \textit{Oxford5k} dataset using our default framework, excluding the time for feature extraction. 
This figure shows that by applying MAX/SUM-mask, our proposed framework can significantly reduce the computational cost, since they help remove about 70\% and 50\% of local conv. features respectively (Section \ref{ssec:effectiveness}).  Additionally, at the dimensionality 512-D, our framework $\mathcal{M}_\text{MAX/SUM} + \phi_\Delta+\psi_\text{d}$ is computationally faster than R-MAC \cite{R-MAC}.


\vspace{\reducevspace}
\section{Conclusion}
\label{sec:conclusion}
In this paper, we present an effective framework which takes activation of convolutional layer as input and produces highly-discriminative image representation for image retrieval. In our proposed framework, we propose to enhance discriminative power of the image representation in two main steps: (i) selecting a representative set of local conv. features using our proposed masking schemes, including SIFT/SUM/MAX mask, then (ii) embedding and aggregating using the state-of-art methods \cite{Temb,FAemb}. Solid experimental results show that the proposed methods compare favorably with the state of the art. A further push the proposed system to achieve very compact binary codes (e.g., by jointly aggregating and hashing \cite{do2017simultaneous} or deep learning-based hashing \cite{do2016learning}) seems interesting future works.

\clearpage
\bibliographystyle{ACM-Reference-Format}
\bibliography{sigprocref} 
\balance

\end{document}